\documentclass[sigconf, screen]{acmart}

\usepackage[utf8]{inputenc} 
\usepackage[T1]{fontenc}    
\usepackage{hyperref}       
\usepackage{url}            
\usepackage{booktabs}       
\usepackage{amsfonts}       
\usepackage{nicefrac}       
\usepackage{microtype}      
\usepackage{pifont}
\usepackage{amsthm}
\usepackage{mathtools}
\usepackage{amsmath}
\usepackage{centernot}
\usepackage{caption}
\usepackage{subcaption}
\usepackage{cleveref}
\usepackage{bbm}
\usepackage{algpseudocode}
\usepackage{afterpage}
\usepackage[suppress]{color-edits}
\usepackage[ruled]{algorithm2e}
\usepackage{graphicx}
\usepackage{ccicons}
\usepackage{tikz}

\newcommand{\indep}{\perp \!\!\! \perp}
\newcommand{\CI}{\mathrel{\perp\mspace{-10mu}\perp}}
\newcommand{\nCI}{\centernot{\CI}}
\newcommand{\xmark}{\ding{53}}%

\DeclareMathOperator*{\argmin}{arg\,min}

\DeclareMathOperator*{\arginf}{arg\,inf}

\SetKwComment{Comment}{$\triangleright$\ }{}

\theoremstyle{definition}
\newtheorem{assumption}{Assumption}
\theoremstyle{plain}
\newtheorem{theorem}{Theorem}[section]
\theoremstyle{plain}

\theoremstyle{definition}

\AtBeginDocument{%
  \providecommand\BibTeX{{%
    \normalfont B\kern-0.5em{\scshape i\kern-0.25em b}\kern-0.8em\TeX}}}

\definecolor{darkred}{rgb}{139, 0, 0}

\addauthor{sw}{blue}
\addauthor{kh}{magenta}
\addauthor{lg}{orange}
\addauthor{ac}{red}

\copyrightyear{2023}
\acmYear{2023}
\setcopyright{acmlicensed}
\acmConference[FAccT '23]{2023 ACM Conference on Fairness, Accountability, and Transparency}{June 12--15, 2023}{Chicago, IL, USA}
\acmBooktitle{2023 ACM Conference on Fairness, Accountability, and Transparency (FAccT '23), June 12--15, 2023, Chicago, IL, USA}
\acmDOI{10.1145/3593013.3594101}
\acmISBN{979-8-4007-0192-4/23/06}

\acmConference[FAccT '23]{}{June 12--15,
  2023}{Chicago, IL}

\acmPrice{}

\begin{document}

\title{Counterfactual Prediction Under Outcome Measurement Error}

\author{Luke Guerdan}
\email{lguerdan@cs.cmu.edu}
\affiliation{%
  \institution{Carnegie Mellon University}
  \city{Pittsburgh}
  \state{PA}
  \country{USA}
}

\author{Amanda Coston}
\email{acoston@cs.cmu.edu}
\affiliation{%
  \institution{Carnegie Mellon University}
  \city{Pittsburgh}
  \state{PA}
  \country{USA}
}

\author{Kenneth Holstein}
\email{kjholste@cs.cmu.edu}
\affiliation{%
  \institution{Carnegie Mellon University}
  \city{Pittsburgh}
  \state{PA}
  \country{USA}
}

\author{Zhiwei Steven Wu}
\email{zstevenwu@cmu.edu}
\affiliation{%
  \institution{Carnegie Mellon University}
  \city{Pittsburgh}
  \state{PA}
  \country{USA}
}

\renewcommand{\shortauthors}{Luke Guerdan, Amanda Coston, Kenneth Holstein, Zhiwei Steven Wu}

\begin{abstract}
Across domains such as medicine, employment, and criminal justice, predictive models often target labels that imperfectly reflect the outcomes of interest to experts and policymakers. For example, clinical risk assessments deployed to inform physician decision-making often predict measures of healthcare utilization (e.g., costs, hospitalization) as a proxy for patient medical need. These proxies can be subject to outcome measurement error when they systematically differ from the target outcome they are intended to measure. However, prior modeling efforts to characterize and mitigate outcome measurement error overlook the fact that the decision being informed by a model often serves as a risk-mitigating intervention that impacts the target outcome of interest and its recorded proxy. Thus, in these settings, addressing measurement error requires counterfactual modeling of treatment effects on outcomes. In this work, we study intersectional threats to model reliability introduced by outcome measurement error, treatment effects, and selection bias from historical decision-making policies. We develop an unbiased risk minimization method which, given knowledge of proxy measurement error properties, corrects for the combined effects of these challenges. We also develop a method for estimating treatment-dependent measurement error parameters when these are unknown in advance. We demonstrate the utility of our approach theoretically and via experiments on real-world data from randomized controlled trials conducted in healthcare and employment domains. As importantly, we demonstrate that models correcting for outcome measurement error or treatment effects alone suffer from considerable reliability limitations. Our work underscores the importance of considering intersectional threats to model validity during the design and evaluation of predictive models for decision support.
\end{abstract}



 

\maketitle

\section{Introduction}\label{sec:introduction}
Algorithmic risk assessment instruments (RAIs) often target labels that imperfectly reflect the goals of experts and policymakers. For example, clinical risk assessments used to inform physician treatment decisions target future utilization of medical resources (e.g., cost, medical diagnoses) as a proxy for patient medical need \citep{obermeyer2019dissecting, mullainathan2021inequity, mullainathan2017does}. Predictive models used to inform personalized learning interventions target student test scores as a proxy for learning outcomes \citep{hur2022using}. Yet, these proxies are subject to \textit{outcome measurement error} (OME) when they systematically differ from the target outcome of interest to domain experts. Unaddressed, OME can be highly consequential: models targeting poor proxies have been linked to misallocation of medical resources \citep{obermeyer2019dissecting}, unwarranted teacher firings \citep{turque2012creative}, and over-policing of minority communities \citep{butcher2022racial}. Given its prevalence and implications, increasing research focus has shifted to understanding and mitigating sources of statistical bias impacting proxy outcomes \citep{de2021leveraging, fogliato2020fairness, fogliato2021validity, wang2021fair, natarajan2013learning, menon2015learning}.

However, prior work modeling outcome measurement error makes a critical assumption that the decision informed by the algorithm does not impact downstream outcomes. Yet this assumption is often unreasonable in decision support applications, where decisions constitute \textit{interventions} that impact the policy-relevant target outcome \textit{and its recorded proxy} \citep{coston2020counterfactual}. For example, in clinical decision support, medical treatments act as risk-mitigating interventions designed to avert adverse health outcomes. However, in the process of selecting a treatment option, a physician will \textit{also} influence measured proxies (e.g., medical cost, disease diagnoses) \citep{obermeyer2019dissecting, mullainathan2021inequity, mullainathan2017does}. As a result, the measurement error characteristics of proxies can vary across the treatment options informed by an algorithm. 

In this work, we develop a counterfactual prediction method that corrects for outcome measurement error, treatment effects, and selection bias in parallel. Our method builds upon \textit{unbiased risk minimization} techniques developed in the label noise literature \citep{natarajan2013learning, patrini2017making, chou2020unbiased, van2015machine}. Given knowledge of measurement error parameters, unbiased risk minimization methods recover an estimator for target outcomes by minimizing a surrogate loss over proxy outcomes. However, existing methods are not designed for \textit{interventional settings} whereby decisions impact outcomes – a limitation that we show severely limits model reliability. Therefore, we develop an unbiased risk minimization technique designed for learning counterfactual models from observational data. We compare our approach against models that correct for OME or treatment effects in isolation by conducting experiments on semi-synthetic data from healthcare and employment domains \citep{finkelstein2012oregon, lalonde1986evaluating, smith2005does}. Results validate the efficacy of our risk minimization approach and underscore the need to carefully vet measurement-related assumptions in consultation with domain experts. Our empirical results also surface systematic model failures introduced by correcting for OME or treatment effects in isolation. To our knowledge, our holistic evaluation is the first to examine how outcome measurement error, treatment effects, and selection bias interact to impact model reliability under controlled conditions.

We provide the following contributions: 1) We derive a problem formulation that models interactions between OME, treatment effects, and selection bias ($\S$ \ref{sec:preliminaries}); 2) We develop a novel approach for learning counterfactual models in the presence of OME ($\S$ \ref{subsec:risk_minimization}). We provide a flexible approach for estimating measurement error rates when these are unknown in advance ($\S$ \ref{subsec:parameter_estimation}); 3) We conduct synthetic and semi-synthetic experiments to validate our approach and highlight reliability issues introduced by modeling OME or treatment effects in isolation ($\S$ \ref{sec:experiments}).

\section{Background and related work}\label{sec:related_work}

\subsection{AI functionality and validity concerns}

Prior work has conducted detailed assessments of specific modeling issues \cite{lakkaraju2017selective, de2021leveraging, kleinberg2018human, coston2020counterfactual, kallus2018residual, wang2021fair}, which have been synthesized into broader critiques of AI validity and functionality \citep{coston2022validity, raji2022fallacy, wang2022against}. \citet{raji2022fallacy} surface AI functionality harms in which models fail to achieve their purported goal due to systematic design, engineering, deployment, and communication failures. \citet{coston2022validity} highlight challenges related to value alignment, reliability, and validity that may draw the justifiability of RAIs into question in some contexts. We build upon this literature by studying \textit{intersectional threats to model reliability} arising from outcome measurement error \citep{jacobs2021measurement, wang2021fair}, treatment effects \citep{coston2020counterfactual, perdomo2020performative}, and selection bias \citep{kallus2018residual} in parallel. 

\begin{figure}
\setlength{\fboxsep}{10pt}
\setlength{\fboxrule}{2pt}
\fcolorbox{gray!50}{gray!5}{%
    \parbox{.85\columnwidth}{%

\textbf{A Motivating Example.} We illustrate the importance of considering interactions between OME and treatment effects by revisiting a widely known audit of an algorithm used to inform screening decisions for a high-risk medical care program  \citep{obermeyer2019dissecting}. This audit surfaced measurement error in a \textit{``cost of medical care''} outcome targeted as a proxy for patient medical need. \textit{Critically, the measurement error analysis performed by \citet{obermeyer2019dissecting} assumes that program enrollment status is independent of downstream cost and medical outcomes.} 

\vspace{3mm}
{\centering
\begin{tabular}{lrrr}
\hline
Sample      &   FPR &   FNR  \\
\hline
Full population &  0.37 &  0.38  \\
Unenrolled      &  0.37 &  0.39  \\
Enrolled        & \textcolor{darkred}{0.64} &  \textcolor{darkred}{0.13}  \\
\hline
\end{tabular}
\par }
\vspace{3mm}

Yet our re-analysis shows that the \textit{``cost of medical care''} proxy has a substantially higher false positive rate and lower false negative rate among program enrollees as compared to the full population (see Appendix \ref{sec:appendix_obermeyer}). This error rate discrepancy is consistent with enrollees receiving closer medical supervision (and as a result, greater costs), even after accounting for their underlying medical need. In this work, we show that failing to model the interactions between OME and treatment effects can introduce substantial model reliability challenges.

        }
    }
\end{figure}

\subsection{Outcome measurement error}

Modeling outcome measurement error is challenging because it introduces two sources of uncertainty: which error model is reasonable for a given proxy, and which specific error parameters govern the relationship between target and proxy outcomes under the \textit{assumed} measurement model \citep{jacobs2021measurement}. Popular error models studied in the machine learning literature include uniform \citep{angluin1988learning, van2015learning}, class-conditional \citep{menon2015learning, scott2013classification}, and instance-dependent \citep{xia2020part, chen2021beyond} structures of outcome misclassification. Work in algorithmic fairness has also studied settings in which measurement error varies across levels of a protected attribute \citep{wang2021fair}, and proposed \lgedit{sensitivity analysis frameworks that are model agnostic}\lgdelete{error model agnostic sensitivity analysis frameworks }\citep{fogliato2020fairness}.

Numerous statistical approaches have been developed for measurement error parameter estimation in the quantitative social sciences literature \citep{roberts1985measurement, bishop1998latent}. Application of these approaches is tightly coupled with domain knowledge of the phenomena under study, as in biostatistics \citep{hui1980estimating} or psychometrics \citep{shrout2012psychometrics}. To date, data-driven techniques for error parameter estimation have primarily been applied in the machine learning literature, which rely on key assumptions relating the target outcome of interest and its proxy \citep{xia2019anchor, liu2015classification, scott2015rate, scott2013classification, menon2015learning, northcutt2021confident}. In this work, we build upon an existing \textit{``anchor assumptions''} framework that estimates error parameters by linking the proxy and target outcome probabilities at specific instances \citep{xia2019anchor}. In contrast to prior work, we provide a range of anchoring assumptions, which can be flexibly combined depending on which are reasonable in a specific algorithmic decision support (ADS) domain. 

\citet{natarajan2013learning} propose a widely-adopted \textit{unbiased risk minimization} approach for learning under noisy labels given knowledge of measurement error parameters \citep{patrini2017making, chou2020unbiased, van2015machine}. This method constructs a surrogate loss $\tilde{\ell}$ such that the $\tilde{\ell}$-risk over proxy outcomes is equivalent to the $\ell$-risk over target outcomes \textit{in expectation}. Additionally, \citet{natarajan2013learning} show that the minimizer of $\tilde{\ell}$-risk over proxy outcomes is optimal with respect to target outcomes if $\ell$ is symmetric (e.g., Huber, logistic, and squared losses). In this work, we develop a novel variant of this unbiased risk minimization approach designed for settings with \textit{treatment-conditional} OME.

 \begin{figure*}[ht]
    \centering
    \includegraphics[width=0.8\textwidth]{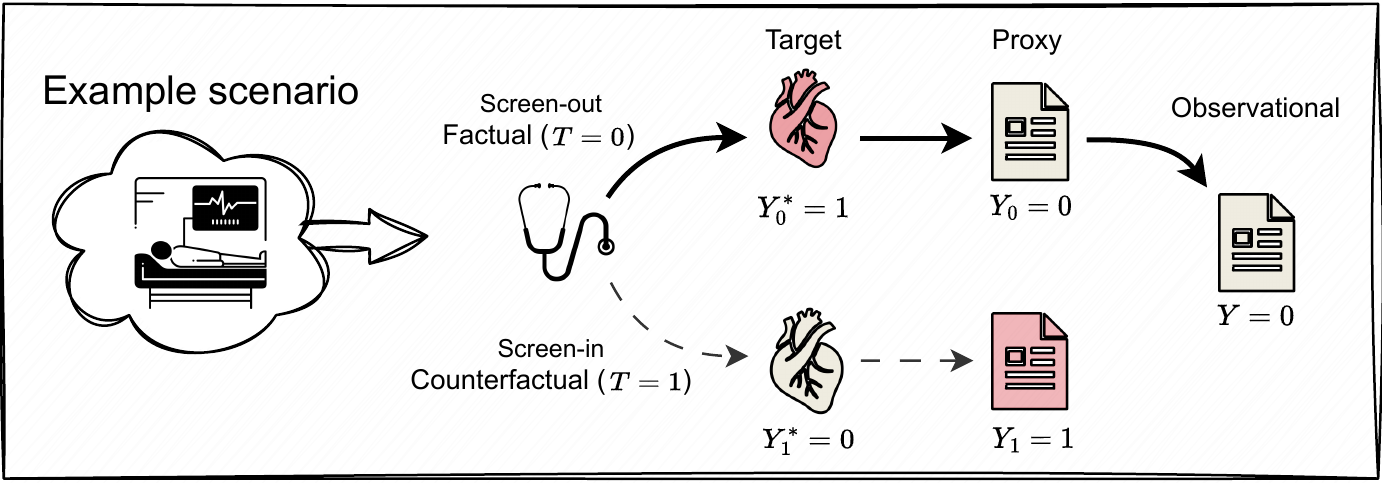}
    \caption{An illustration of treatment-conditional OME in heart attack prediction. Under the factual decision to screen-out from a high-risk care management program ($T=0$), heart attack occurred ($Y^*_0=1$) but went undiagnosed ($Y_0=0$). Under the counterfactual decision to screen in ($T=1$), heart attack \textit{would have} been averted ($Y^*_1=0$) but would have been incorrectly diagnosed ($Y_1=1$). The observed outcome in medical records reflects the proxy value under factual decision to screen-out ($Y=0$).}\label{fig:tc_ome}
\end{figure*}

\subsection{Counterfactual prediction}

Recent work has shown that counterfactual modeling is necessary when the decision informed by a predictive model serves as a risk-mitigating intervention \citep{coston2020counterfactual}. Building off of this result, we argue that it is necessary to account for treatment effects on \textit{target outcomes of interest and their observed proxy} while modeling OME. Our methods build upon conditional average treatment effect (CATE) estimation techniques from the causal inference literature \citep{shalit2017estimating, johansson2020generalization, abrevaya2015estimating}. Subject to identification conditions \citep{pearl2009causal, rubin2005causal}, these approaches predict the difference between the expected outcome under treatment (e.g., high-risk program enrollment) versus control (e.g., no program enrollment) conditional on covariates. One family of \textit{outcome regression estimators} predicts the CATE by directly estimating the expected outcome under treatment or control conditional on covariates \citep{hill2011bayesian, kunzel2019metalearners, chernozhukov2017double}. However, these methods suffer from statistical bias when prior decisions were non-randomized (i.e., due to distribution shift induced by selection bias) \citep{shimodaira2000improving, angluin1988learning}. Therefore, we leverage a re-weighting strategy proposed by \citep{johansson2020generalization} to correct for this selection bias during risk minimization. Our re-weighting method performs a similar bias correction as inverse probability weighting (IPW) methods \citep{shimodaira2000improving, shimodaira2000improving, rosenbaum1983central}.

Outcome measurement error has also been studied in causal inference literature. \citet{finkelstein2021partial} bound the average treatment effect (ATE) under multiple plausible OME models. \citet{shu2019causal} propose a doubly robust method which accounts for measurement error during ATE estimation, while \citet{diaz2013sensitivity} provide a sensitivity analysis framework for examining robustness of ATE estimates to OME. This work is primarily concerned with estimating \textit{population statistics} rather than predicting outcomes conditional on measured covariates (i.e., the CATE).

\section{Preliminaries}\label{sec:preliminaries}

Let $p^*(X, T, Y^*_0, Y^*_1, Y_0, Y_1)$ be a fixed joint distribution over covariates $X \in \mathcal{X} \subseteq \mathbb{R}^{d}$, past decisions\footnote{We also use the word \textit{treatments} to refer to binary decisions. This draws upon historical applications of causal inference to medical settings.} $T \in \{ 0, 1 \}$, \textit{target} potential outcomes $\{Y^*_0, Y^*_1\}  \in \mathcal{Y} \subseteq \{ 0, 1 \}$, and \textit{proxy} potential outcomes $\{Y_0, Y_1\}  \in \mathcal{Y} \subseteq \{ 0, 1 \}$. Under the potential outcomes framework \citep{rubin2005causal}, $\{Y^*_0, Y_0\}$ and  $\{Y^*_1, Y_1\}$ are the target and proxy outcomes that \textit{would occur} under $T=0$ and $T=1$, respectively (Figure \ref{fig:tc_ome}). Building upon the class-conditional model studied in observational settings \cite{natarajan2013learning, menon2015learning}, we propose a treatment-conditional outcome measurement error model, whereby the class probability of the proxy potential outcome is given by 
\begin{equation}\label{eq:tce_model}
\eta_t(x)=(1-\beta_t) \cdot \eta^*_t(x) + \alpha_t \cdot (1 - \eta^*_t(x)),\;\; \forall x \in X
\end{equation}
where $\alpha_t \coloneqq p(Y_t=1 \mid Y_t^*=0 )$, $\beta_t \coloneqq p(Y_t=0 \mid Y_t^*=1)$ are the proxy false positive and false negative rates under treatment $t \in \{0, 1 \}$ such that  $\alpha_t + \beta_t < 1$. This model imposes the following assumption on the structure of measurement error. 

\begin{assumption}[Measurement error] 
Measurement error rates are fixed across covariates: $Y \CI X \mid  Y^*\lgedit{, T}$. \label{assumption:error_model} \accomment{According to the causal diagram in Fig 2, there's a path from X to Y through T right?}
\end{assumption}

While we make this assumption to foreground study of treatment effects, our methods are also compatible with approaches designed for error rates that vary across covariates \citep{wang2021fair} (see $\S$ \ref{subsec:measurement_model_discussion} for discussion). Given the joint $p^*$, we would like to estimate $\eta^*_t(x) \coloneqq p(Y^*_t = 1 \mid X=x)$, for any target covariates $x \in X$, which is the 
probability of the target potential outcome under intervention $t \in \{ 0, 1\}$. However, rather than observing $Y^*_t$ directly, we sample from an \textit{observational distribution} $p(X, T, Y)$, where $Y \in  \mathcal{Y} \subseteq \{0, 1 \}$ is an observed \textit{proxy outcome}. By consistency, the \lgedit{unobserved target potential outcome and} observed proxy potential outcome is determined by the treatment assignment.

\begin{assumption}[Consistency] $ \lgedit{Y^* = T \cdot Y^*_1 + (1-T) \cdot Y^*_0;} \; Y = T \cdot Y_1 + (1-T) \cdot Y_0$.
\label{assumption:consistency}
\end{assumption}

\lgedit{This assumption holds that the target and proxy potential outcomes ${Y^*_t, Y_t}$ are observed among instances assigned to treatment $t$ \citep{pearl2009causal, rubin2005causal, rubin1974estimating}.} To identify \lgedit{observational proxy outcomes} $Y$, we also require the following additional causal assumptions.

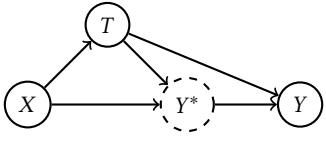
\begin{figure}[t]
    \centering
    \begin{tikzpicture}[node distance={15mm}, thick, main/.style = {draw, circle}] 
    \node[circle, draw] (1) {$X$}; 
    \node[circle, draw]  (2) [above right of=1] {$T$}; 
    \node[circle, draw, dashed] (4) [ below right of=2] {$Y^*$}; 
    \node[circle, draw] (3) [right of=4] {$Y$}; 
    \draw[->] (1) -- (2); 
    \draw[->] (2) -- (4); 
    \draw[->] (4) -- (3); 
    \draw[->] (1) -- (4); 
    \draw[->] (2) -- (3); 
    \end{tikzpicture}
    \caption{A causal diagram of treatment-conditional outcome measurement error.}
    \label{fig:dag}
\end{figure}

\begin{assumption}[Ignorability] $\{ \lgedit{Y^*_0, Y^*_1}, Y_0, Y_1\} \indep T \mid X$. \lgdelete{This holds that no unmeasured confounders jointly influence decisions and proxy or target potential outcomes.}\lgedit{This holds that target and proxy potential outcomes are unconfounded given measured covariates $X$.}
\label{assumption:ignorability}
\end{assumption}
Ignorability can be violated in decision support applications when unobservables impact both the treatment and outcome \citep{kleinberg2018human, lakkaraju2017selective, de2021leveraging}. Understanding and addressing limitations introduced by ignorability is a major ongoing research focus \citep{diaz2013sensitivity, rambachan2022counterfactual, coston2020counterfactual}. We provide follow-up discussion of this assumption in $\S$ \ref{sec:discussion_causal}.

\begin{assumption}[Positivity] $\forall x \in X,\; 0 > p(T=1|X=x) > 1$. This holds that each instance $x \in X$ has some chance of receiving each decision $t \in \{0, 1 \}$.\label{assumption:positivity}
\end{assumption}

 Positivity is often reasonable in decision support applications because instances $x \in X$ that require support from predictive models are subject to discretionary judgement due to uncertainty. Instances that are certain to receive a given treatment (i.e., $p(T=1|X=x)=0$ or $p(T=1|X=x)=1$) would normally be routed via a different administrative procedure. Figure \ref{fig:dag} shows a causal diagram representing the data generating process we study in this work.

\section{Methodology}\label{sec:methodology}

We begin by developing an unbiased risk minimization approach which recovers an estimator for $\eta^*_t$ given knowledge of error parameters ($\S$ \ref{subsec:risk_minimization}). We then provide a method for estimating $\alpha_t$ and $\beta_t$ when error parameters are unknown in advance ($\S$ \ref{subsec:parameter_estimation}). 

\subsection{Unbiased risk minimization}\label{subsec:risk_minimization}

In this section, we develop an approach for estimating $\eta^*_t$ given observational data drawn from $p(X, T, Y)$ and measurement error parameters $\alpha_t$, $\beta_t$. Let $f_t\in\mathcal{H}$ for $\mathcal{H} \subset\{f_t: \mathcal{X}  \rightarrow [ 0,1 ]\}$ be a probabilistic decision function targeting $Y^*_t$ and let $\ell: \mathcal{Y} \times [0, 1] \rightarrow \mathbb{R}_+$ be a loss function. If we \lgedit{observed} target potential outcomes $Y^*_t \sim p^*$, we \lgedit{could} directly apply supervised learning techniques to minimize the expected $\ell$-risk of $f_t$ over target potential outcomes 
\begin{equation}\label{eq:erm_target}
R_{\ell}^*(f_t) \coloneqq  \mathbb{E}_{p^*} [ \ell(f_t(X), Y^*_t)]
\end{equation}
and learn an estimator for $\eta^*_t$ via standard empirical risk minimization approaches. \lgedit{Given a \textit{strongly proper composite} loss such} that\ $\argmin_{f_t} R_{\ell}^*(f_t)$ is a monotone transform $\psi$ of $\eta^*_t$ (e.g., the logistic and exponential loss), \lgedit{this would enable recovering} class probabilities from the optimal prediction via the link function $\psi$ \citep{agarwal2014surrogate, menon2015learning}. However, directly minimizing (\ref{eq:erm_target}) is not possible in our setting because we sample observational proxies instead of target potential outcomes. We address this challenge by constructing a \textit{re-weighted surrogate risk} $R_{t,\tilde{\ell}}^{w}$ such that evaluating this risk over observed proxy outcomes is equivalent to $R_{\ell}^*$ in expectation.

In particular, let $w:\mathcal{X} \rightarrow \mathbb{R}_+$ be a weighting function satisfying $\mathbb{E}_{X}[w(X)|T=t] = 1$ and let $\ell: \mathcal{Y} \times [0, 1] \rightarrow \mathbb{R}_+$ be a surrogate loss function. We construct a \textit{re-weighted surrogate risk}
\begin{equation}\label{eq:rwrm_sl}
R_{t,\tilde{\ell}}^{w}\left(f_t\right):=\mathbb{E}_{p}\left[w(X) \tilde{\ell}(f_t(X), Y) \mid T=t\right]
\end{equation}

such that $R_{\ell}^*(f_t) = R_{t, \tilde{\ell}}(f_t)$ in expectation. Theorem \ref{thm:unbiased_risk} shows that we can recover a surrogate risk satisfying this property by constructing $w(x)$ as in (\ref{eq:re-weighting}) and $\tilde{\ell}$ as in (\ref{eq:surrogate_loss}). Note that this surrogate risk requires knowledge of $\alpha_t$, $\beta_t$.

\begin{theorem}\label{thm:unbiased_risk}

Assume treatment-conditional error (\ref{assumption:error_model}), consistency (\ref{assumption:consistency}), ignorability (\ref{assumption:ignorability}) and positivity (\ref{assumption:positivity}). Then under target intervention $t \in \{ 0, 1\}$, $R^*_{\ell}(f_t)=R_{t, \tilde{\ell}}^w(f_t)$ for the weighting function $w: \mathcal{X} \rightarrow \mathbb{R}_+$ given by
\begin{equation}\label{eq:re-weighting}
w(x) \coloneqq \frac{p(T=t)}{(2 t-1)\cdot \pi(x)+1-t} 
\end{equation}

and surrogate loss $\tilde{\ell}: \mathcal{Y} \times [0, 1] \rightarrow \mathbb{R}_+$  given by
\begin{equation}\label{eq:surrogate_loss}
\begin{aligned}
\tilde{\ell}(f_t(x), 1) \coloneqq \frac{(1-\alpha_t) \cdot \ell(f_t(x), 1) - \beta_t  \cdot \ell(f_t(x), 0)}{1-\beta_t-\alpha_t}\\
\tilde{\ell}(f_t(x), 0) \coloneqq \frac{(1-\beta_t)  \cdot \ell(f_t(x), 0) - \alpha_t  \cdot \ell(f_t(x), 1)}{1-\beta_t-\alpha_t}
\end{aligned}
\end{equation}
where in (\ref {eq:re-weighting}), $\pi(x) \coloneqq p(T=1|X=x)$ is the propensity score function.

\end{theorem}

We prove Theorem \ref{thm:unbiased_risk} in Appendix \ref{sec:appendix_proof}. Intuitively, $R_{t,\tilde{\ell}}^{w}\left(f_t\right)$ applies a \textit{joint bias correction} for OME and distribution shift introduced by historical decision-making policies (i.e., selection bias). The unbiased risk minimization framework dating back to \citet{natarajan2013learning} corrects for OME by minimizing a surrogate loss $\tilde{\ell}$ on proxies $Y$ observed \textit{over the full population unconditional on treatment}. Yet this approach is untenable when decisions impact outcomes ($T \nCI \{Y^*, Y\} $) and error rates differ across treatments. One possible extension of unbiased risk minimizers to the treatment-conditional setting involves minimizing $\tilde{\ell}$ over the treatment population $p(X|T=t)$
\begin{equation}\label{eq:rwrm_sl}
R_{t,\tilde{\ell}}\left(f_t\right):=\mathbb{E}_{p}\left[\tilde{\ell}(f_t(X), Y) \mid T=t\right].
\end{equation}

However, $R_{t,\tilde{\ell}} \neq R_{\ell}^*$ in observational settings because the treatment population $p(X|T=t)$ can differ from the marginal population $p(X)$ under historical selection policies when $X \nCI T$. Therefore, our re-weighting procedure applies a second bias correction that adjusts $p(X|T=t)$ to resemble $p(X)$.

\textit{Learning algorithm}. As a result of Theorem \ref{thm:unbiased_risk}, we can learn a predictor $\hat{\eta}^*_t$ by minimizing the re-weighted surrogate risk over \textit{observed samples} $(X_1, T_1, Y_1), ..., (X_n, T_n, Y_n)  \sim p$.  First, we estimate the weighting function $\hat{w}(x)$ through a finite sample, which boils down to learning propensity scores $\hat{\pi}(x)$ (as shown in \eqref{eq:re-weighting}). Estimating the propensity scores can be done by applying supervised learning algorithms to learn a predictor from $X$ to $T$. Then for any treatment $t$, weighting function $\hat w$, and predictor $f_t$, we can approximate $R_{t,\tilde{\ell}}^{w}\left(f_t\right)$ by taking the sample average over the treatment population
\begin{equation}\label{eq:rwsl}
\hat{R}^{\hat{w}}_{t, \tilde{\ell}}(f_t) \coloneqq \frac{1}{n_t} \sum_{i : T_i = t} \hat{w}(X_i)\tilde{\ell}(f_t(X_i), Y_i)
\end{equation}
for $n_t = \sum_{i=1}^{n}\mathbbm{1}[T_i=t]$. Therefore, given $\hat{w}$ we can learn a predictor from observational data by minimizing the empirical risk 
\begin{equation}\label{eq:argmin_rwsl}
\hat{f_t} \leftarrow \argmin_{f_t \in \mathcal{H}} \hat{R}^{\hat{w}}_{t, \tilde{\ell}}(f_t).   
\end{equation}
We refer to solving (\ref{eq:argmin_rwsl}) as \textit{re-weighted risk minimization with a surrogate loss} (Algorithm \ref{alg:rwsl}). 

\begin{algorithm}[h]
    \SetKwInput{Input}{Input}
    \SetKwInput{Output}{Output}
    \caption{Re-weighted risk minimization with surrogate loss (RW-SL)}\label{alg:rwsl}
    \Input{Data $\mathcal{W} = \{(X_i, T_i, Y_i)\}_{i=1}^n \sim p$}
    \Output{Learned estimator $\hat{\eta}^*_t(x)$ }
    Partition $\mathcal{W}$ into $\mathcal{W}_1$, $\mathcal{W}_2$, $\mathcal{W}_3$ \\
    On $\mathcal{W}_1$, estimate parameters $\hat{\alpha}_t, \hat{\beta}_t \leftarrow \text{CCPE}(\mathcal{W}_1)$\\
    On $\mathcal{W}_2$, learn $\hat{\pi}(x)$ by regressing $T \sim X$ \\
    On $\mathcal{W}_3$, use $\hat{\pi}(x), \hat{\alpha}_t, \hat{\beta}_t$ to solve $\hat{\eta}_t^*(x) \leftarrow \argmin_{f_t \in \mathcal{H}} \hat{R}^{\hat{w}}_{t, \tilde{\ell}}(f_t)$ \\
\end{algorithm}

\begin{algorithm}[h]
    \SetKwInput{Input}{Input}
    \SetKwInput{Output}{Output}
    \caption{Conditional class probability estimation (CCPE)}\label{alg:ccpe}
    \Input{Data $\mathcal{V} \sim p$}
    \Output{Parameter estimates $\hat{\alpha}_t$, $\hat{\beta}_t$}
    Partition $\mathcal{V}$ into $\mathcal{V}_1$, $\mathcal{V}_2$ \\
    On $\mathcal{V}_1$, learn $\hat{\eta}_t(x)$ by regressing $Y \sim X \; | \; T = t$\\
    On $\mathcal{V}_2$, estimate error parameters: $\hat{\alpha}_t  = \underset{{x \in X}}{\min} \{ \hat{\eta}_t(x) \},\;\; \hat{\beta}_t = 1 - \underset{{x \in X}}{\max}  \{ \hat{\eta}_t(x) \} $
\end{algorithm}

\subsection{Error parameter identification and estimation}\label{subsec:parameter_estimation}

Because our risk minimization approach requires knowledge of OME parameters, we develop a method for estimating $\alpha_t$, $\beta_t$ from observational data. Error parameter estimation is challenging in decision support applications because target outcomes often result from nuanced social and organizational processes. Understanding the measurement error properties of proxies targeted in criminal justice, medicine, and hiring domains remains an ongoing focus of domain-specific research \citep{fogliato2021validity, akpinar2021effect, mullainathan2021inequity, obermeyer2019dissecting, zwaan2015challenges, chalfin2016productivity}. \textit{Therefore, we develop an approach compatible with \textit{multiple sources of domain knowledge about proxies}, which can be flexibly combined depending on which assumptions are deemed reasonable in a specific context.} 

Error parameters are \textit{identifiable} if they can be uniquely computed from observational data. Because our error model (e.q. \ref{eq:tce_model}) expresses the proxy class probability as a linear equation with two unknowns, $\alpha_t$, $\beta_t$ are identifiable if the target class probability $c^*_{t,i} = \eta^*_t(x_i)$ and proxy class probability $c_{t, i} = \eta_t(x_i)$ are known at two distinct points $(c^*_{t,i}, c_{t, i})$ and $(c^*_{t,j}, c_{t, j})$ such that $c^*_{t,i} \neq c^*_{t,j}$. Following prior literature \citep{han2020survey}, we refer to knowledge of $(c^*_{t,i}, c_{t, i})$ as an \textit{anchor assumption} because it requires knowledge of the unobserved quantity $\eta^*_t$. We now introduce several anchor assumptions that are practical in ADS, before showing that these can be flexibly combined to identify $\alpha_t$, $\beta_t$ in Theorem \ref{theorem:anchor_identification}.

\textbf{Min anchor.} A min anchor assumption holds if there is an instance at no risk of the target potential outcome under intervention $t$: $c_{t, i}^* = \inf_{x_i \in \mathcal{X}} { \{ \eta^*_t(x_i) \} } = 0$. Because $\eta_t$ is a strictly monotone increasing transform of $\eta^*_t$, the corresponding value of $\eta_t$ can be recovered via $c_{t, i} = \inf_{x_i \in \mathcal{X}} \{ \eta_t(x_i) \}$ \citep{menon2015learning}. Min anchors are reasonable when there are cases that are confirmed to be at no risk based on domain knowledge of the data generating process. For example, a min anchor may be reasonable in diagnostic testing if a patient is confirmed to be negative for a medical condition based on a high-precision gold standard medical test \citep{enoe2000estimation}. 

\textbf{Max anchor.} A max anchor assumption holds if there is an instance at certain risk of the target outcome under intervention $t$:  $c_{t, i}^* = \sup_{x_i \in \mathcal{X}} \{ \eta^*_t(x_i) \} = 1$. The corresponding value of $\eta_t$ can be recovered via $c_{t, i} = \sup_{x_i \in \mathcal{X}} \{ \eta_t(x_i) \}$ because $\eta_t$ is a strictly monotone increasing transform of $\eta^*_t$. Max anchors are reasonable when there are confirmed instances of a positive target potential outcome based on domain knowledge of the data generating process. For example, a max anchor may be justified in a medical setting if a subset of patients have confirmed disease diagnoses based on biopsy results \citep{begg1983assessment}, or if a disease prognosis (and resulting health outcomes) are known from pathology. 

\textbf{Base rate anchor.} A base rate anchor assumption holds if the expected value of $\eta^*_t$ is known under intervention $t$: $c_{t, i}^* = \mathbb{E}[\eta^*_t(X)]$. The corresponding value of $\eta_t$ can be recovered by taking the expectation over the proxy class probability $c_{t, i} = \mathbb{E}[\eta_t(X)]$. Base rate anchors are practical because the prevalence of unobservable target outcomes (e.g., medical conditions \citep{walter1988estimation}, crime \citep{kruttschnitt2014estimating, lohr2019measuring}, student performance \citep{schouwenburg2004procrastination, di2022epidemiology}) is routinely estimated via domain-specific analyses of measurement error. For instance, studies have been conducted to estimate the base rate of undiagnosed heart attacks (i.e., accounting for measurement error in diagnosis proxy outcomes) \cite{orso2017epidemiology}. Additionally, the conditional average treatment effect $\mathbb{E}[\eta^*_1(X)] - \mathbb{E}[\eta^*_0(X)]$ is commonly estimated in randomized controlled trials (RCTs) while assessing treatment effect heterogeneity \citep{hill2011bayesian}. While the conditional average treatment effect is normally estimated via proxies $Y_0$ and $Y_1$, measurement error analysis is a routine component of RCT design and evaluation \citep{gamerman2019pragmatic}. 

Anchor assumptions can be flexibly combined to identify error parameters based on which set of assumptions are reasonable in a given ADS domain. In particular, Theorem \ref{theorem:anchor_identification} shows that combinations of anchor assumptions listed in Table \ref{fig:identification_table} are sufficient for identifying error parameters under our causal assumptions. 

\begin{table}[t]
\centering
\begin{tabular}{@{}lccccc@{}}

                          & \textbf{Know $\alpha_t$} & \textbf{Min} & \textbf{Base rate} & \textbf{Max} & \textbf{Know $\beta_t$} \\ 
                          \cmidrule(l){2-6}
\textbf{Know $\alpha_t$}  &      \xmark              &     \xmark         &    \checkmark       &      \checkmark        &   \checkmark  \\
\textbf{Min}              &       \xmark             &      \xmark         &      \checkmark       &     \checkmark        &   \checkmark  \\
\textbf{Base rate}        &       \checkmark          &    \checkmark  &       \xmark              &   \checkmark           &   \checkmark  \\ 
\textbf{Max}              &      \checkmark         &   \checkmark     &   \checkmark         &   \xmark  &       \xmark          \\
\textbf{Know $\beta_t$}   &      \checkmark         &   \checkmark   &  \checkmark          &  \xmark   &      \xmark          
\end{tabular}
\caption{Multiple combinations of min, max, and base rate anchor assumptions (shown via $\checkmark$) enable identification of $\alpha_t$, $\beta_t$.}\label{fig:identification_table}

\vspace{-3mm}
\end{table}

\begin{theorem}\label{theorem:anchor_identification}
Assume treatment-conditional error (\ref{assumption:error_model}),  consistency (\ref{assumption:consistency}), ignorability (\ref{assumption:ignorability}) and positivity (\ref{assumption:positivity}). Then $\alpha_t, \beta_t$ are identifiable from observational data $p(X, T, Y)$ given any identifying pair of anchor assumptions provided in Table \ref{fig:identification_table}.
\end{theorem}

We \lgedit{prove} Theorem \ref{theorem:anchor_identification} in Appendix \ref{sec:appendix_proof}. In practice, we estimate the error rates on finite samples $(X_i, T_i, Y_i) \sim p$, which gives an approximation $\hat{\eta}_t$. Therefore, we propose a conditional class probability estimation (CCPE) method for parameter estimation which estimates $\hat{\alpha}_t$, $\hat{\beta}_t$ by fitting $\hat{\eta}_t$ on observational data then applying the relevant pair of anchor assumptions to estimate error rates. Algorithm \ref{alg:ccpe} provides pseudocode for this approach with min and max anchors, which can easily be extended to other pairs of identifying assumptions shown in Table \ref{fig:identification_table}. The combination of min and max anchors is known as \textit{weak separability} \citep{menon2015learning} or \textit{mutual irreducibility} \citep{scott2013classification, scott2015rate} in the observational label noise literature. Prior results in the observational setting show that unconditional class probability estimation (i.e., fitting $\hat{\eta}(x)=p(Y=1|X=x$)) yields a consistent estimator for observational error rates under weak seperability \citep{scott2013classification, reeve2019classification}. Statistical consistency results extend to the treatment-conditional setting under positivity (\ref{assumption:positivity}) because $p(T=t|X=x) > 0, \; \forall t \in \{0, 1\}, \; x \in \mathcal{X}$. However, asymptotic convergence rates may be slower under strong selection bias if $p(T=t|X=x)$ is near 0. 

\section{Experiments}\label{sec:experiments}
Experimental evaluation under treatment-conditional OME is challenging due to compounding sources of uncertainty. We do not observe counterfactual outcomes in historical data, making it challenging to estimate the quality of new models via observational data. Further, because the target outcome is not observed directly, we rely on measurement assumptions when studying proxy outcomes in naturalistic data. We address this challenge by conducting a controlled evaluation with synthetic data where ground truth potential outcomes are fully observed. To better reflect the ecological settings of real-world deployments, we also conduct a semi-synthetic evaluation with real data collected through randomized controlled trials (RCTs) in healthcare and employment domains. Our evaluation (1) validates our proposed risk minimization approach, (2) underscores the need to carefully consider measurement assumptions during error rate estimation, and (3) shows that correcting for OME or treatment effects in isolation is insufficient.\footnote{\lgedit{Code for all experiments can be found at: \url{https://github.com/lguerdan/CP_OME}.}}

\subsection{Models}\label{subsec:baselines}

We compare several modeling approaches in our evaluation to examine how existing modeling practices are impacted by treatment-conditional outcome measurement error: 
 
 \begin{itemize}

     \item \textbf{Unconditional proxy (UP)}. Predict the observed outcome unconditional on treatment: $X \rightarrow Y$. \lgdelete{This model does not adjust for OME or treatment effects, and describes performance in a setting in which a practitioner does not correct for any of the challenges discussed in this work.} \lgdelete{Reflecting current practice for deployed models, }This model \textit{does not adjust for OME or treatment effects.}\lgedit{, and reflects model performance in a scenario in which practitioners overlook all challenges examined in this work. }\footnote{This baseline is also called an \textit{observational risk assessment} in experiments reported by \citet{coston2020counterfactual}.}
     
     \item \textbf{Unconditional target (UT)}. Predict the target outcome unconditional on treatment: $X \rightarrow Y^*$. Here, we determine $Y^*$ by applying consistency: $Y^* = (1-T) \cdot Y^*_0 + T \cdot Y^*_1$. This method reflects a setting in which no OME is present but modeling does not account for treatment effects \citep{wang2021fair, natarajan2013learning, menon2015learning, obermeyer2019dissecting}.

     \item \textbf{Conditional proxy \textbf{(CP}).} Predict the proxy outcome conditional on treatment: $X, T \rightarrow Y$. This is a counterfactual model that estimates a conditional expectation \textit{without correcting for OME} \citep{coston2020counterfactual, shalit2017estimating, kunzel2019metalearners}.\footnote{This model is known by different names in the causal inference literature, including the backdoor adjustment (G-computation) formula \citep{pearl2009causal, robins1986new}, T-learner \cite{kunzel2019metalearners}, and plug-in estimator \citep{kennedy2022semiparametric}.}

     \item \textbf{Re-weighted surrogate loss (RW-SL).} Our proposed risk minimization approach, as defined in equation (\ref{eq:argmin_rwsl}). This method corrects for both OME and treatment effects in parallel. \lgedit{Additionally, this method corrects for distribution shift due to selection bias in the prior decision-making policy via re-weighting.} 

     \item \textbf{Target Potential Outcome (TPO).} Directly predict the target potential outcome: $X \rightarrow Y^*_t$. This model is an \textit{oracle} that provides an upper-bound on model performance under no OME or treatment effects. 
     
 \end{itemize}
 
We also perform an ablation of our proposed RW-SL method by including a model that applies a surrogate loss correction $\tilde{\ell}$ over the treatment population without re-weighting (\textbf{SL}).

\begin{figure*}[htp]%
    \centering
    \begin{minipage}{0.24\textwidth}
        \includegraphics[width=\textwidth, trim={0 2mm 0 0},clip]{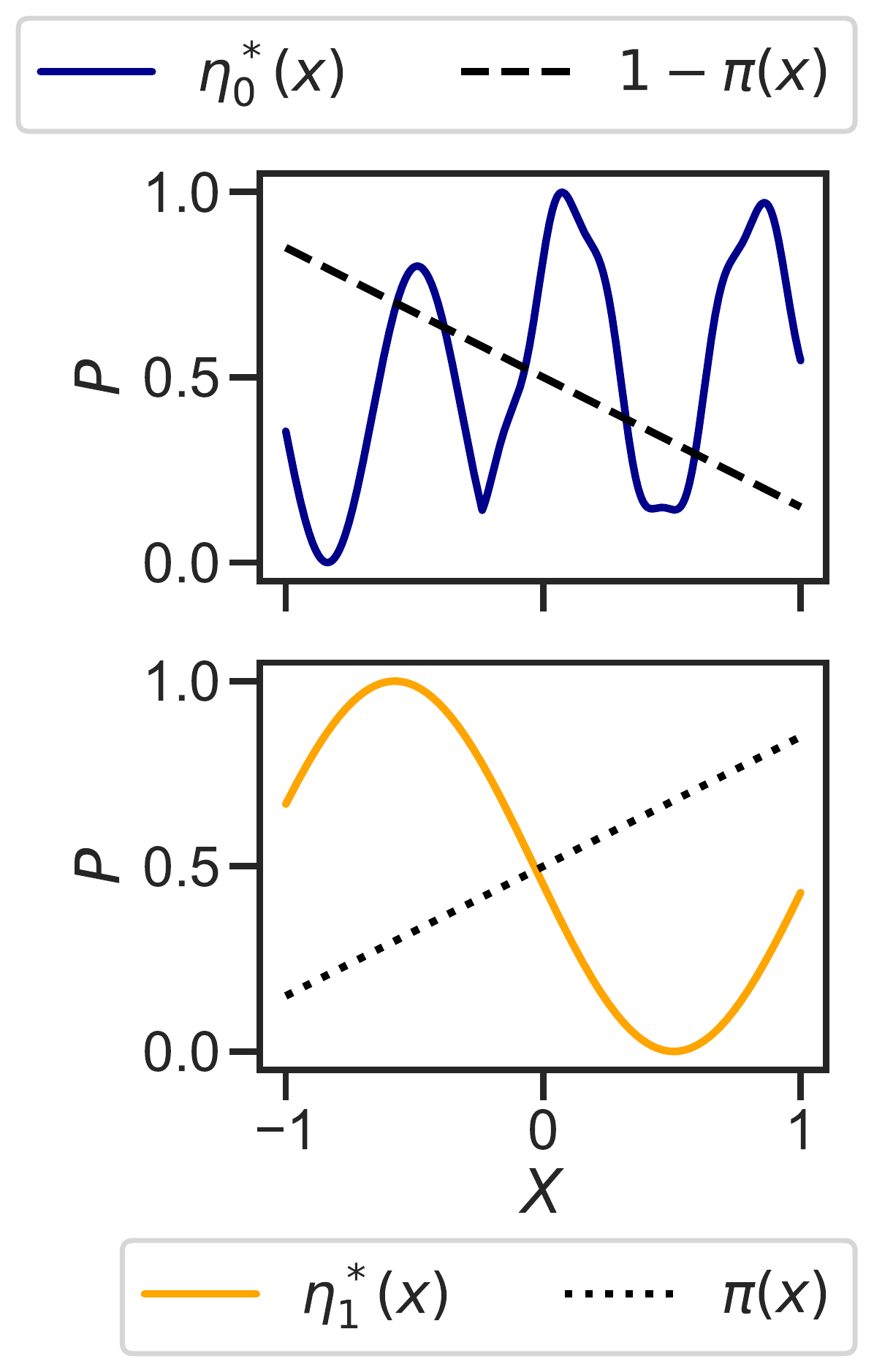}
        \caption{Synthetic setup.}
        \label{fig:syn_setup}
    \end{minipage}\hfill
    \begin{minipage}{0.7\textwidth}
\includegraphics[width=\textwidth, trim={0 2mm 0 0},clip]{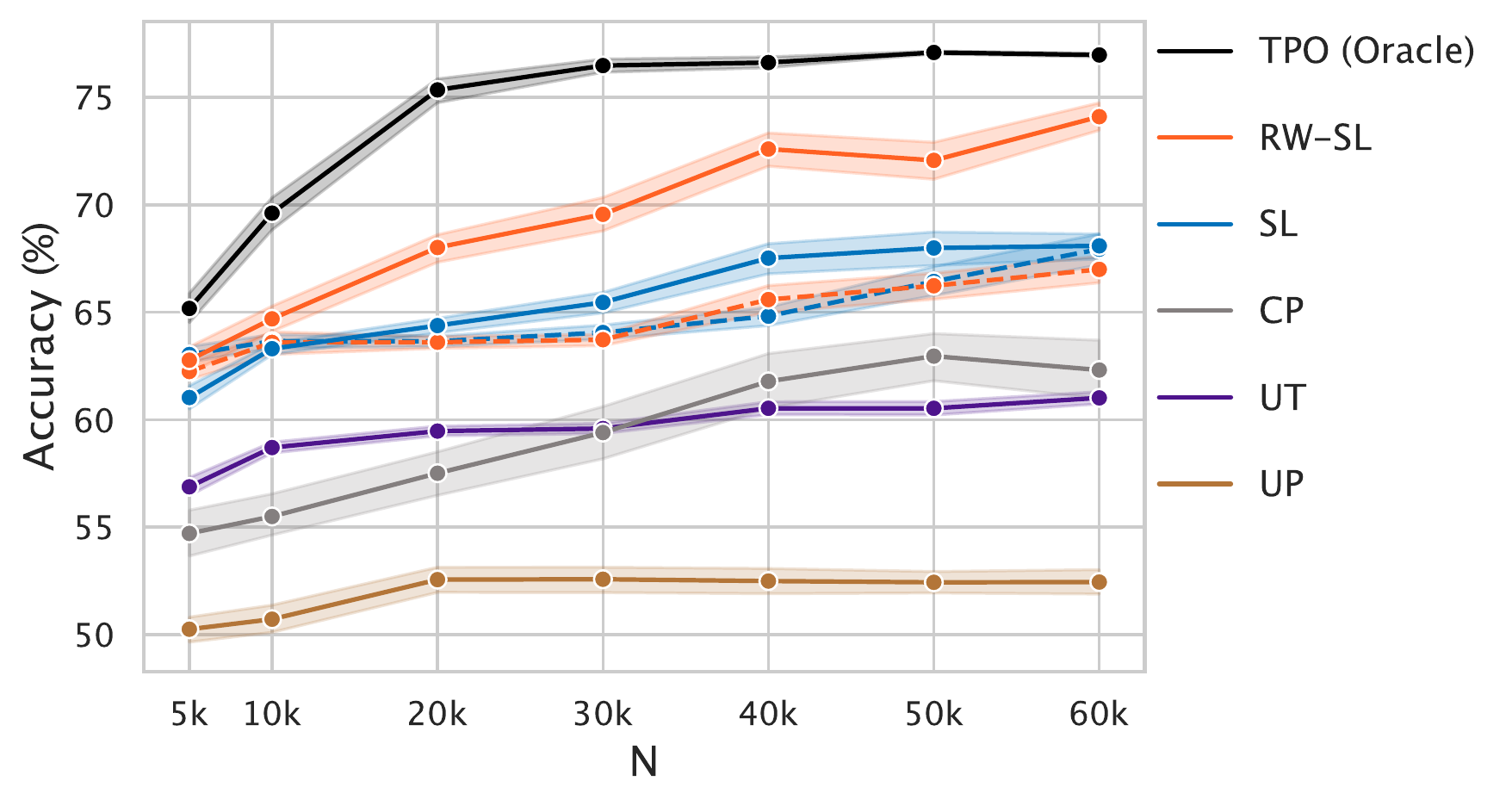}
\caption{Accuracy on $Y^*_0$ as a function of sample size. RW-SL and SL with oracle parameters plotted with solid lines. RW-SL and SL with learned parameters plotted with dashed lines. Results averaged over asymmetric error settings reported in Table \ref{table:syn_results}. }\label{fig:syn_convergence}
\end{minipage}\par
\vskip\floatsep

\begin{center}
{\renewcommand{\arraystretch}{1.1}%
\begin{tabular}{lccccc}
\toprule
$(\alpha_0, \beta_0)$   &     (0.0, 0.4) &     (0.1, 0.3) &     (0.2, 0.2) &     (0.3, 0.1) &     (0.4, 0.0) \\
\midrule
UP    &  54.18 (0.09) &  53.00 (0.39) &  54.89 (1.09) &  55.81 (0.74) &  46.76 (0.33) \\
UT       &  61.57 (0.63) &  60.95 (0.50) &  60.49 (0.41) &  61.00 (0.49) &  60.54 (0.70) \\
CP              &  51.36 (1.83) &  68.24 (2.61) &  \textbf{75.05} (0.92) &  67.77 (1.33) &  61.88 (0.28) \\
\hline
SL $(\hat{\alpha}, \hat{\beta})$ &  72.38 (1.65) &  65.45 (0.66) &  67.43 (1.64) &  68.01 (0.99) &  65.92 (1.34) \\
RW-SL $(\hat{\alpha}, \hat{\beta})$  &  69.08 (1.55) &  65.96 (1.18) &  66.57 (1.32) &  68.39 (1.33) &  64.56 (0.52) \\
SL $(\alpha, \beta)$     &  \lgedit{67.09 (1.24)} &  \lgedit{67.58 (1.19)} &  \lgedit{67.75 (1.08)} &  \lgedit{69.11 (1.17)} &  \lgedit{68.59 (1.41)} \\
RW-SL $(\alpha, \beta)$      &  \textbf{73.68} (1.49) &  \textbf{73.39} (1.60) &  72.52 (1.66) &  \textbf{74.34} (1.15) &  \textbf{75.01} (1.24) \\
\hline
TPO    &  \textbf{77.08} (0.11) &  \textbf{77.09} (0.20) &  \textbf{76.98} (0.08) &  \textbf{76.84} (0.18) &  \textbf{76.90} (0.16) \\
\bottomrule
\end{tabular}
}
\end{center}
\captionof{table}{Model accuracy (s.e.) across error parameter settings $(\alpha_0, \beta_0)$ at $N=60k$ samples over 10 runs. Top-2 performance across each ($\alpha_0$, $\beta_0$) setting shown in bold.}\label{table:syn_results}

\end{figure*}

\subsection{Experiments on synthetic data}

We begin by experimentally manipulating treatment effects and measurement error via a synthetic evaluation. Because this provides full control over the data generating process, we can evaluate methods against target potential outcomes. This evaluation would not possible with real-world data because counterfactual target outcomes are unobserved. Our experiment design is consistent with prior synthetic evaluations of counterfactual risk assessments \cite{coston2020counterfactual} and causal inference methods \citep{shi2019adapting, nie2021vcnet}. In our evaluation, we sample outcomes via the following data generating process:

\begin{enumerate}
     \item $Y^*_t := \;\; \sim \text{Bern}(\eta_t^*(X)), \; \forall t \in \{ 0, 1\}$ 
    \item  $Y_t := \begin{cases}
              1-\epsilon_+ & \text{if $Y^*_t=1$, where $\epsilon_+ \sim \text{Bern}(\beta_t)$ } \\
              \epsilon^-  & \text{if $Y^*_t=0$, where $\epsilon_- \sim \text{Bern}(\alpha_t)$ } \\
            \end{cases}, \forall t \in \{0, 1\} \;  
        $
    \item $T := \;\;  \sim \text{Bern}(\pi(X))$
     \item  $Y^* := (1-T) \cdot Y^*_0 + T \cdot Y^*_1; \;  Y := (1-T) \cdot Y_0 + T \cdot Y_1$
    
 \end{enumerate}
As shown in Figure \ref{fig:syn_setup}, we draw  $X \sim U(-1, 1)$ and sample target potential outcomes from sinusoidal class probability functions (see Appendix \ref{sec:appendix_experiments} for details). Note that our choice of $\eta^*_0(x)$, $\eta^*_1(x)$ satisfies min and max anchor assumptions. Because $\eta^*_0(x)$ and $\eta^*_1(x)$ differ, models that do not condition on treatment (i.e., UP, UT) will learn an average of the two class probability functions. Under our choice of $\pi(x)$, fewer samples are drawn from $\eta^*_1(x)$ in the region where $\pi(x)$ is small (near $x=-1$), and fewer samples are drawn from $\eta^*_0(x)$ in the region where $1-\pi(x)$ is small (near $x=1$). This introduces selection bias when sampling from $\pi(x)$.

\subsubsection{Setup details} We train each model in $\S$ \ref{subsec:baselines} to predict risk under no intervention ($t=0$) and vary $(\alpha_0, \beta_0)$. We keep $(\alpha_1, \beta_1)$ fixed at $(0,0)$ across settings. When estimating OME parameters, we run CCPE with \lgedit{sample splitting and} cross-fitting (Algorithm \ref{alg:ccpe_crossfit}) with min and max anchor assumptions for identification. These assumptions hold precisely under this controlled evaluation (Figure \ref{fig:syn_setup}). We run all methods with \lgedit{sample splitting and} cross-fitting (Algorithm \ref{alg:rw_sl_crossfit}) and report performance on $Y^*_0$.

\subsubsection{Results} Figure \ref{fig:syn_convergence} shows the performance of each model as a function of sample size. TPO provides an upper bound on performance because it learns directly from target potential outcomes. RW-SL with oracle parameters $(\alpha, \beta)$ outperforms all other methods trained on observational data \lgedit{across across the full range of sample sizes. Thus, while Thm. \ref{thm:unbiased_risk} shows that RW-SL recovers an unbiased risk estimator \textit{in expectation}, this method also demonstrates favorable finite-sample performance characteristics in practice. This finding is inline with prior experimental evaluations of unbiased risk estimators reported in the standard supervised learning setting \citep{natarajan2013learning, wang2021fair}, and is further supported by reliable performance characteristics we observe in small sample regimes (see Appendix \ref{sec:appendix_experiments}).} 

\lgedit{In contrast,} both models that do not condition on treatment (UP and UT), and the conditional regression trained on proxy outcomes (CP), reach a performance plateau by 50k samples and do not benefit from additional data. This indicates that (1) learning a counterfactual model and (2) correcting for measurement error is necessary to learn $\eta^*_t$ in this evaluation. We likely observe a sharper plateau in UP and UT above 20k samples because these approaches fit a weighted average of $\eta^*_0$ and $\eta^*_1$ (where $\eta^*_1$ differs from $\eta^*_0$ considerably). \lgedit{We observe that RW-SL and SL performance deteriorates with learned parameters $(\hat{\alpha}, \hat{\beta})$ across all sample size settings due to misspecification in learned parameter estimates and weights.}

Table \ref{table:syn_results} shows a breakdown across error rates $(\alpha_0, \beta_0)$ at $60k$ samples. RW-SL outperforms SL when oracle parameters are known. However, RW-SL and SL perform comparably when weights and parameters are learned. This may be because RW-SL relies on estimates $\hat{w}$ in addition to $\hat{\alpha}_0, \hat{\beta}_0$, which could introduce instability given misspecification in $\hat{w}$. CP performs notably well under high error parameter symmetry (i.e., $\alpha_0=\beta_0=.2$). This is consistent with prior results from the class-conditional label noise literature \citep{natarajan2013learning, menon2015learning}, which show that the optimal classifier threshold for misclassification risk does not change under symmetric label noise. CP performs worse under high error asymmetry. \lgedit{We do not observe a similar performance improvement in UP and UT in the symmetric error setting because these baselines learn a weighted combination of $\eta_0$ and $\eta_1$, which differs from the target function $\eta^*_0$ at all classification thresholds.}

\subsection{Semi-synthetic experiments on healthcare and employment data}

In addition to our synthetic evaluation, we conduct experiments using real-world data collected as part of randomized controlled trials (RCTs) in healthcare and employment domains. While this evaluation affords less control over the data generating process, it provides a more realistic sample of data likely to be encountered in real-world model deployments. Evaluation via data from randomized or partially randomized experimental studies is useful for validating counterfactual prediction approaches because random assignment ensures that causal assumptions are satisfied \citep{shalit2017estimating, johansson2020generalization, coston2020runtime}.

\subsubsection{Randomized Controlled Trial (RCT) Datasets} In 2008, the U.S. state of Oregon expanded access to its Medicare program via a lottery system \citep{finkelstein2012oregon}. This lottery provided an opportunity to study the effects of Medicare enrollment on healthcare utilization and medical outcomes via an experimental design, commonly referred to as the Oregon Health Insurance Experiment (OHIE). Lottery enrollees completed a pre-randomization survey recording demographic factors and baseline health status and were given a one-year follow-up assessment of health status and medical care utilization. We refer the reader to \citet{finkelstein2012oregon} for details. We use the OHIE dataset to construct an evaluation task that parallels the label choice bias analysis of \citet{obermeyer2019dissecting}. We use this dataset rather than synthetic data released by \citet{obermeyer2019dissecting} because (1) treatment was randomly assigned, ruling out positivity and ignorability violations possible in observational data, and (2) OHIE data contains covariates necessary for predictive modeling. We predict diagnosis with an active chronic medical condition over the one-year follow-up period given $D=58$ covariates, including health history, prior emergency room visits, and public services use. We predict chronic health conditions because findings from \citet{obermeyer2019dissecting} indicate that this outcome variable is a reasonable choice of proxy for patient medical need. We adopt the randomized lottery draw as the treatment. \footnote{The OHIE experiment had imperfect compliance ($\approx$ 30 percent of selected individuals successfully enrolled \citep{finkelstein2012oregon}). Therefore, we predict diagnosis with a new chronic health condition given the \textit{opportunity to enroll} in Medicare. This evaluation is consistent with many high-stakes decision-support settings granting opportunities to individuals, which they have a choice to pursue if desired.} 

We conduct a second real-world evaluation using the JOBS dataset, which investigates the effect of job retraining on employment status \citep{shalit2017estimating}. This dataset includes an experimental sample collected by \citet{lalonde1986evaluating} via the National Supported Work (NSW) program (297 treated, 425 control) consisting primarily of low-income individuals seeking job retraining. \citet{smith2005does} combine this sample with a ``PSID'' comparison group (2,490 control) collected from the general population, which resulted in a final sample with 297 treated and 2,915 control. This dataset includes $D=17$ covariates including age, education, prior earnings, and interaction terms. 482 (15\%) of subjects were unemployed at the end of the study. Following \citet{johansson2020generalization}, we construct an evaluation task predicting unemployment under enrollment ($t=1$) and no enrollment ($t=0$) in a job retraining program conditional on covariates.

\begin{figure*}[t]
    \centering
    \includegraphics[width=\textwidth]{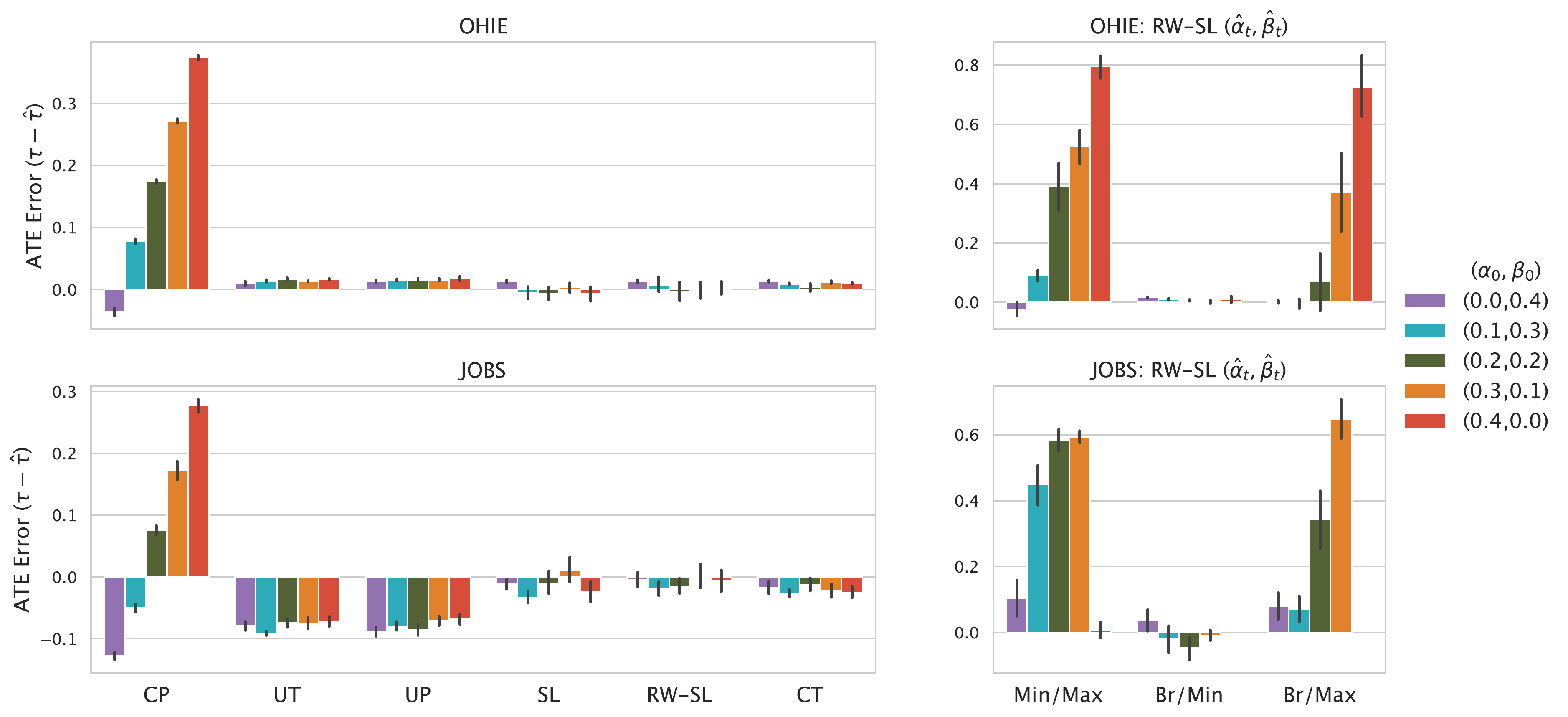}
    \caption{Bias in ATE estimates on OHIE and JOBS data. Error bars indicate standard error over ten runs. CT is a model with oracle access to target outcomes and RW-SL is our proposed approach. }\label{fig:ate_error}
\end{figure*}

\subsubsection{Synthetic OME and selection bias}\label{subsec:syn_ome_selection} We experimentally manipulate OME to examine how outcome regressions perform under treatment-conditional error of known magnitude. We adopt diagnosis with a new chronic condition and future unemployment as a \textit{target outcome} for OHIE and JOBS, respectively. We observe proxy outcomes by flipping target outcomes with probability ($\alpha_0$, $\beta_0$). We keep ($\alpha_1$, $\beta_1$) fixed at $(0, 0)$. This procedure of generating proxy outcomes by flipping available labels is a common approach for vetting the feasibility of new methodologies designed to address OME \citep{wang2021fair, menon2015learning, natarajan2013learning}. This approach offers precise control over the magnitude of OME but suffers from less ecological validity than studying multiple \lgedit{naturalistic} proxies \citep{obermeyer2019dissecting}. We opt for this semi-synthetic evaluation because (1) the precise measurement relationship between naturally occurring proxies may not be fully known, (2) the measurement relationship between naturally occurring proxies cannot be manipulated experimentally, and (3) there are few RCT datasets (i.e., required to guarantee causal assumptions) that contain multiple proxies of the same target outcome.

Models used for decision support are typically trained using data gathered under a historical decision-making policy. When prior decisions were made non-randomly, this introduces selection bias ($T \nCI X$) and causes distribution shift between the population that received treatment $t$ in training data, and the full population encountered at deployment time. Therefore, we emulate selection bias in the \textit{training dataset}, and evaluate models over a held-out test set of randomized data. We insert selection bias in OHIE data by removing individuals from the treatment (lottery winning) arm with household income above the federal poverty line (10\% of the treatment sample). This resembles an observational setting in which low-income individuals are more likely to receive an opportunity to enroll in a health insurance program (e.g., Medicaid, which determines eligibility based on household income in relation to the federal poverty line). We restrict our analysis to single-person households, yielding $N=12,994$ total samples ($6,189$ treatment, $6,805$ control). 

We model selection bias in JOBS data by including samples from the observational and experimental cohorts in the training data. Because the PSID comparison group consists of individuals with higher income and education than the NSW group, there is considerable distribution shift across the NSW and PSID cohorts \citep{lalonde1986evaluating, smith2005does, johansson2020generalization}. Therefore, a model predicting unemployment over the control population (consisting of NSW and PSID samples) may suffer from bias when evaluated against test data that only includes samples from the NSW experimental arm. Thus we split data from the NSW experimental cohort 50-50 across training and test dataset, and only include PSID data in the training dataset.

\subsubsection{Experimental setup}

We include a Conditional Target (CT) model in place of a TPO model because counterfactual outcomes are not available in experimental data. CT provides a reasonable upper-bound on performance because identifiability conditions are satisfied in an experimental setting \citep{pearl2009causal}. However, it is not possible to report accuracy over potential outcomes because counterfactual outcomes are unobserved. Therefore, we report error in ATE estimates $\tau - \hat{\tau}$, for 
$$
\tau := \mathbb{E}[Y^* \mid T=1] - \mathbb{E}[Y^* \mid T=0], \;\;\; \hat{\tau} := \mathbb{E}[\hat{\eta}_1(X)] -  \mathbb{E}[\hat{\eta}_0(X)] 
$$
where \lgedit{$\tau$ corresponds to the ground-truth treatment effect reported by prior work \citep{johansson2020generalization, denteh2022increases} and} $\hat{\eta}_t$ is a learned model discussed in $\S$ \ref{subsec:baselines}. One subtlety of this comparison is that our outcome regressions target the \textit{conditional} average treatment effect, while $\tau$ reflects the ATE across the full population. Following prior evaluations \citep{johansson2020generalization}, we compare all methods against the ATE because the ground-truth CATE is not available for JOBS or OHIE data.  \footnote{ \lgedit{While our insertion of synthetic selection bias (\S \ref{subsec:syn_ome_selection}) introduces distribution shift such that $p(X|T=1)$ differs from $p(X|T=0)$, it does not alter ground-truth $\tau$ because the conditional outcome distribution $p(Y^*|T)$ remains unchanged. This setup recreates the unconfounded observational setting in which causal identification assumptions are satisfied \citep{rubin1974estimating}.}} We report results over a test fold of randomized data that does not contain flipped outcomes or selection bias. Appendix \ref{sec:appendix_experiments} contains additional setup details.

\subsubsection{Results} Figure \ref{fig:ate_error} shows bias in ATE estimates $\tau - \hat{\tau}$ over 10 runs on JOBS and OHIE data. The left panel compares CP, UT, UP, and the oracle CT model against RW-SL/SL with oracle parameters $(\alpha_0, \beta_0)$, $(\alpha_1, \beta_1)$. We show performance of RW-SL with learned parameters $(\hat{\alpha}_0, \hat{\beta}_0)$, $(\hat{\alpha}_1, \hat{\beta}_1)$ on the right panel. The left panel shows that CP is highly sensitive to measurement error. This is because measurement error introduces bias in estimates of the conditional expectations, which propagates to treatment effect estimates. Because UT  and UP  do not condition on treatment, they estimate an \textit{average} of the outcome functions $\eta_0^*$ and $\eta_1^*$, and generate predictions near 0. Therefore, while UT   and UP  perform well on OHIE data due to a small ground-truth ATE ($\tau = 0.015$), they perform poorly on JOBS ($\tau = -0.077$). SL and RW-SL with oracle parameters $\alpha_t$, $\beta_t$ perform comparably to the CT model with oracle access to target outcomes across all measurement error settings. 

\lgedit{While we observe that re-weighting improves performance in our synthetic evaluation (given oracle parameters), we do not observe a similar advantage of RW-SL over SL in this experiment. Our results parallel other empirical evaluations of re-weighting for counterfactual modeling tasks on real-world data (e.g., see $\S$ 3.4.2 in \citep{coston2020counterfactual}). One potential explanation for this finding is that \acedit{our predictive model class (multi-layer MLPs) is large enough to learn} the target regressions $\eta^*_0$ and $\eta^*_1$ \acdelete{are easy to fit on} \acedit{for } OHIE and JOBS data, even after our insertion of synthetic selection bias. As a result, re-weighting may not be required to learn a reasonable estimate of $\eta^*_0$ and $\eta^*_1$ given available data. This interpretation is supported by strong performance of the oracle CT model. \acdelete{Because we fit a multilayer MLP for the outcome regressions, model misspecification is not a likely explanation for our findings \citep{sugiyama2007covariate}.} } \accomment{I made the "our model class is flexible point" earlier so I commented out this sentence, but feel free to revert/slightly change if this is making a point about misspecification beyond what I have now.}

As shown on the right panel of Figure \ref{fig:ate_error}, RW-SL performance is highly sensitive to the choice of anchor assumption used to estimate parameters ($\hat{\alpha}_0$, $\hat{\beta}_0$), ($\hat{\alpha}_1$, $\hat{\beta}_1$) as indicated by increased bias in $\hat{\tau}$ and greater variability over runs. In particular, RW-SL performs poorly when Min/Max and Br/Max pairs of anchor assumptions are used to estimate error rates because the max anchor assumption is violated on OHIE and JOBS data. \lgedit{We shed further light on this finding by fitting the CT model to estimate $\hat{\eta}^*_0$, $\hat{\eta}^*_1$ on OHIE data, then computing inferences over a validation fold $X_{val}$. This analysis reveals that}
\begin{align*}
    \min_{x \in X_{val}} \hat{\eta}^*_0 \approx 2.23 \cdot e^{-6} &,\;\; \max_{x \in X_{val}} \hat{\eta}^*_0 \approx 0.85 \\
    \min_{x \in X_{val}} \hat{\eta}^*_1 \approx 1.02 \cdot e^{-5} &,\;\;  \max_{x \in X_{val}} \cdot \hat{\eta}^*_1 \approx 0.81
\end{align*}
which suggests that the min anchor assumption that $\min_{x\in X_{val}} \hat{\eta}^*_t = 0$ is reasonable for $t \in \{ 0, 1 \}$, while the max anchor assumption that $\max_{x\in X_{val}} \hat{\eta}^*_t = 1$ is violated for both $t \in \{ 0, 1 \}$. Therefore, because the min anchor assumption is satisfied for these choices of target outcome, and the ground-truth base rate is known in this experimental setting, RW-SL demonstrates strong performance \lgedit{given the} Br/Min combination of anchor assumptions. \lgedit{In contrast, because the max anchor is violated, estimating $\beta_t$ by taking the supremium of $\hat{\eta}_t(x)$ introduces bias in $\hat{\beta}_t$, which results in poor performance of RW-SL with Min/Max and Br/Max anchors.} Applying this same procedure to the unemployment outcome targeted in JOBS data also reveals a violation of the max anchor assumption. These results underscore the importance of selecting anchor assumptions in close consultation with domain experts because it is not possible to verify anchor assumptions by learning $\hat{\eta}^*_t$ when the target outcome of interest is unobserved.

\section{Discussion}\label{sec:discussion}
In this work, we show the importance of carefully addressing intersectional threats to model reliability during the development and evaluation of predictive models for decision support. Our theoretical and empirical results validate the efficacy of our unbiased risk minimization approach. When OME parameters are known, our method performs comparably to a model with oracle access to target potential outcomes. However, our results underscore the importance of vetting anchoring assumptions used for error parameter estimation before using error rate estimates for risk minimization. Critically, our experimental results also demonstrate that correcting for a single threat to model reliability in isolation is insufficient to address model validity concerns \citep{raji2022fallacy}, and risks promoting false confidence in corrected models. Below, we expand upon key considerations surfaced by our work. 

\subsection{Decision points and complexities in measurement error modeling}\label{subsec:measurement_model_discussion}

Our work speaks to key complexities faced by domain experts, model developers, and other stakeholders while examining proxies in ADS. One decision surfaced by our work entails which \textit{measurement error model} best describes the relationship between the unobserved outcome of policy interest and its recorded proxy. We open this work by highlighting a measurement model decision made by \citet{obermeyer2019dissecting} during their audit of a clinical risk assessment: that error rates are fixed across treatments. Our work suggests that failing to account for treatment-conditional error in OME models may exacerbate reliability concerns. However, at the same time, the error model we adopt in this work intentionally abstracts over other factors known to impact proxies in decision support tasks, including error rates that vary across covariates. Although this simplifying assumption can be unreasonable in some settings \citep{fogliato2021validity,akpinar2021effect}, including the one studied by \citet{obermeyer2019dissecting}, it is helpful in foregrounding previously-overlooked challenges involving treatment effects and selection bias. In practice, model developers correcting for measurement error may wish to combine our methods with existing unbiased risk minimization approaches designed for group-dependent error where appropriate \citep{wang2021fair}. Further, analyses of measurement error should not overlook more fundamental conceptual differences between target outcomes and proxies readily available for modeling (e.g., when long-term child welfare related outcomes targeted by a risk assessment differ from \textit{immediate} threats to child safety weighted by social workers \citep{kawakami2022improving, kawakami2022improving}). This underscores the need to carefully weigh the validity of proxies in consultation with multiple stakeholders (e.g., domain experts, data scientists, and decision-makers) while deciding whether OME correction is warranted. 

A second decision point highlighted in this work entails the \textit{specific measurement error parameters} that govern the relationship between target and proxy outcomes. In particular, our work underscores the need for a tighter coupling between domain expertise and data-driven approaches for error parameter estimation. Current techniques designed to address OME in the machine learning literature -- which typically examine settings with ``label noise'' -- rely heavily upon data-driven approaches without close consideration of whether the underlying measurement assumptions hold \citep{wang2021fair, northcutt2021confident, menon2015learning, xia2019anchor}. While application of these assumptions may be practical for methodological evaluations and theoretical analysis \citep{scott2013classification, scott2015rate, reeve2019classification}, these assumptions should be carefully vetted when applying OME correction to real-world proxies. This is supported by our findings in Figure \ref{fig:ate_error}, which show that RW-SL performs poorly when the anchor assumptions used for error parameter estimation are violated. Our flexible set of anchor assumptions provides a step towards a tighter coupling between domain expertise and data-driven approaches in measurement parameter estimation.

\subsection{Challenges of linking causal and statistical estimands}\label{sec:discussion_causal}

Our counterfactual modeling approach requires several causal identifiability assumptions \citep{pearl2009causal}, which may not be satisfied in all decision support contexts. Of our assumptions, the most stringent is likely ignorability, which requires that no unobserved confounders influenced past decisions and outcomes. While recent modeling developments may ease ignorability-related concerns in some cases \cite{coston2020counterfactual, rambachan2022counterfactual}, model developers should carefully evaluate whether confounders are likely to impact a model in a given deployment context. At the same time, our results show that formulating algorithmic decision support as a \textit{``pure prediction problem''} that optimizes predictive performance without estimating causal effects \citep{kleinberg2015prediction} imposes equally serious limitations. If the policy-relevant target outcome of interest is risk \textit{conditional on intervention} (as is often the case in decision support applications), an observational model will generate invalid predictions for cases that historically responded most to treatment \citep{coston2020counterfactual}. Our results, which empirically demonstrate poor performance of observational PU and TU models that overlook treatment-effects, corroborate prior findings indicating that counterfactual modeling is required to ensure the reliability of RAIs in decision support settings \citep{coston2020counterfactual}. Taken together, our work suggests that domain experts and model developers should exercise considerable caution while mapping the causal estimand of policy interest to the statistical estimand targeted by a predictive model \citep{lundberg2021your}.

\begin{acks}\label{sec:acknowledgements} 
We thank \lgedit{the anonymous reviewers and} attendees of the NeurIPS 2022 Workshop on Causality for Real-world Impact for their helpful feedback. This work was supported by an award from the UL Research Institutes through the Center for Advancing Safety of Machine Intelligence (CASMI) at Northwestern University, the Carnegie Mellon University Block Center for Technology and Society (Award No. 53680.1.5007718), and the National Science Foundation Graduate Research Fellowship Program (Award No. DGE-1745016). \lgedit{ZSW is supported in part by an NSF FAI Award (No. 1939606), a Google Faculty Research Award, a J.P. Morgan Faculty Award, a Facebook Research Award, an Okawa Foundation Research Grant, and a Mozilla Research Grant.}
\end{acks}

\bibliographystyle{ACM-Reference-Format}
\bibliography{refs}


\begin{thebibliography}{80}


\ifx \showCODEN    \undefined \def \showCODEN     #1{\unskip}     \fi
\ifx \showDOI      \undefined \def \showDOI       #1{#1}\fi
\ifx \showISBNx    \undefined \def \showISBNx     #1{\unskip}     \fi
\ifx \showISBNxiii \undefined \def \showISBNxiii  #1{\unskip}     \fi
\ifx \showISSN     \undefined \def \showISSN      #1{\unskip}     \fi
\ifx \showLCCN     \undefined \def \showLCCN      #1{\unskip}     \fi
\ifx \shownote     \undefined \def \shownote      #1{#1}          \fi
\ifx \showarticletitle \undefined \def \showarticletitle #1{#1}   \fi
\ifx \showURL      \undefined \def \showURL       {\relax}        \fi
\providecommand\bibfield[2]{#2}
\providecommand\bibinfo[2]{#2}
\providecommand\natexlab[1]{#1}
\providecommand\showeprint[2][]{arXiv:#2}

\bibitem[Abrevaya et~al\mbox{.}(2015)]%
        {abrevaya2015estimating}
\bibfield{author}{\bibinfo{person}{Jason Abrevaya}, \bibinfo{person}{Yu-Chin
  Hsu}, {and} \bibinfo{person}{Robert~P Lieli}.}
  \bibinfo{year}{2015}\natexlab{}.
\newblock \showarticletitle{Estimating conditional average treatment effects}.
\newblock \bibinfo{journal}{\emph{Journal of Business \& Economic Statistics}}
  \bibinfo{volume}{33}, \bibinfo{number}{4} (\bibinfo{year}{2015}),
  \bibinfo{pages}{485--505}.
\newblock


\bibitem[Agarwal(2014)]%
        {agarwal2014surrogate}
\bibfield{author}{\bibinfo{person}{Shivani Agarwal}.}
  \bibinfo{year}{2014}\natexlab{}.
\newblock \showarticletitle{Surrogate regret bounds for bipartite ranking via
  strongly proper losses}.
\newblock \bibinfo{journal}{\emph{The Journal of Machine Learning Research}}
  \bibinfo{volume}{15}, \bibinfo{number}{1} (\bibinfo{year}{2014}),
  \bibinfo{pages}{1653--1674}.
\newblock


\bibitem[Akpinar et~al\mbox{.}(2021)]%
        {akpinar2021effect}
\bibfield{author}{\bibinfo{person}{Nil-Jana Akpinar}, \bibinfo{person}{Maria
  De-Arteaga}, {and} \bibinfo{person}{Alexandra Chouldechova}.}
  \bibinfo{year}{2021}\natexlab{}.
\newblock \showarticletitle{The effect of differential victim crime reporting
  on predictive policing systems}. In \bibinfo{booktitle}{\emph{Proceedings of
  the 2021 ACM conference on fairness, accountability, and transparency}}.
  \bibinfo{pages}{838--849}.
\newblock


\bibitem[Angluin and Laird(1988)]%
        {angluin1988learning}
\bibfield{author}{\bibinfo{person}{Dana Angluin} {and} \bibinfo{person}{Philip
  Laird}.} \bibinfo{year}{1988}\natexlab{}.
\newblock \showarticletitle{Learning from noisy examples}.
\newblock \bibinfo{journal}{\emph{Machine Learning}} \bibinfo{volume}{2},
  \bibinfo{number}{4} (\bibinfo{year}{1988}), \bibinfo{pages}{343--370}.
\newblock


\bibitem[Begg and Greenes(1983)]%
        {begg1983assessment}
\bibfield{author}{\bibinfo{person}{Colin~B Begg} {and}
  \bibinfo{person}{Robert~A Greenes}.} \bibinfo{year}{1983}\natexlab{}.
\newblock \showarticletitle{Assessment of diagnostic tests when disease
  verification is subject to selection bias}.
\newblock \bibinfo{journal}{\emph{Biometrics}} (\bibinfo{year}{1983}),
  \bibinfo{pages}{207--215}.
\newblock


\bibitem[Bishop(1998)]%
        {bishop1998latent}
\bibfield{author}{\bibinfo{person}{Christopher~M Bishop}.}
  \bibinfo{year}{1998}\natexlab{}.
\newblock \showarticletitle{Latent Variable Models.}
\newblock \bibinfo{journal}{\emph{Learning in graphical models}}
  \bibinfo{volume}{371} (\bibinfo{year}{1998}).
\newblock


\bibitem[Butcher et~al\mbox{.}(2022)]%
        {butcher2022racial}
\bibfield{author}{\bibinfo{person}{Bradley Butcher}, \bibinfo{person}{Chris
  Robinson}, \bibinfo{person}{Miri Zilka}, \bibinfo{person}{Riccardo Fogliato},
  \bibinfo{person}{Carolyn Ashurst}, {and} \bibinfo{person}{Adrian Weller}.}
  \bibinfo{year}{2022}\natexlab{}.
\newblock \showarticletitle{Racial Disparities in the Enforcement of Marijuana
  Violations in the US}. In \bibinfo{booktitle}{\emph{Proceedings of the 2022
  AAAI/ACM Conference on AI, Ethics, and Society}}. \bibinfo{pages}{130--143}.
\newblock


\bibitem[Chalfin et~al\mbox{.}(2016)]%
        {chalfin2016productivity}
\bibfield{author}{\bibinfo{person}{Aaron Chalfin}, \bibinfo{person}{Oren
  Danieli}, \bibinfo{person}{Andrew Hillis}, \bibinfo{person}{Zubin Jelveh},
  \bibinfo{person}{Michael Luca}, \bibinfo{person}{Jens Ludwig}, {and}
  \bibinfo{person}{Sendhil Mullainathan}.} \bibinfo{year}{2016}\natexlab{}.
\newblock \showarticletitle{Productivity and selection of human capital with
  machine learning}.
\newblock \bibinfo{journal}{\emph{American Economic Review}}
  \bibinfo{volume}{106}, \bibinfo{number}{5} (\bibinfo{year}{2016}),
  \bibinfo{pages}{124--127}.
\newblock


\bibitem[Chen et~al\mbox{.}(2021)]%
        {chen2021beyond}
\bibfield{author}{\bibinfo{person}{Pengfei Chen}, \bibinfo{person}{Junjie Ye},
  \bibinfo{person}{Guangyong Chen}, \bibinfo{person}{Jingwei Zhao}, {and}
  \bibinfo{person}{Pheng-Ann Heng}.} \bibinfo{year}{2021}\natexlab{}.
\newblock \showarticletitle{Beyond class-conditional assumption: A primary
  attempt to combat instance-dependent label noise}. In
  \bibinfo{booktitle}{\emph{Proceedings of the AAAI Conference on Artificial
  Intelligence}}, Vol.~\bibinfo{volume}{35}. \bibinfo{pages}{11442--11450}.
\newblock


\bibitem[Chernozhukov et~al\mbox{.}(2017)]%
        {chernozhukov2017double}
\bibfield{author}{\bibinfo{person}{Victor Chernozhukov}, \bibinfo{person}{Denis
  Chetverikov}, \bibinfo{person}{Mert Demirer}, \bibinfo{person}{Esther Duflo},
  \bibinfo{person}{Christian Hansen}, {and} \bibinfo{person}{Whitney Newey}.}
  \bibinfo{year}{2017}\natexlab{}.
\newblock \showarticletitle{Double/debiased/neyman machine learning of
  treatment effects}.
\newblock \bibinfo{journal}{\emph{American Economic Review}}
  \bibinfo{volume}{107}, \bibinfo{number}{5} (\bibinfo{year}{2017}),
  \bibinfo{pages}{261--265}.
\newblock


\bibitem[Chou et~al\mbox{.}(2020)]%
        {chou2020unbiased}
\bibfield{author}{\bibinfo{person}{Yu-Ting Chou}, \bibinfo{person}{Gang Niu},
  \bibinfo{person}{Hsuan-Tien Lin}, {and} \bibinfo{person}{Masashi Sugiyama}.}
  \bibinfo{year}{2020}\natexlab{}.
\newblock \showarticletitle{Unbiased risk estimators can mislead: A case study
  of learning with complementary labels}. In
  \bibinfo{booktitle}{\emph{International Conference on Machine Learning}}.
  PMLR, \bibinfo{pages}{1929--1938}.
\newblock


\bibitem[Coston et~al\mbox{.}(2020a)]%
        {coston2020runtime}
\bibfield{author}{\bibinfo{person}{Amanda Coston}, \bibinfo{person}{Edward
  Kennedy}, {and} \bibinfo{person}{Alexandra Chouldechova}.}
  \bibinfo{year}{2020}\natexlab{a}.
\newblock \showarticletitle{Counterfactual predictions under runtime
  confounding}.
\newblock \bibinfo{journal}{\emph{Advances in Neural Information Processing
  Systems}}  \bibinfo{volume}{33} (\bibinfo{year}{2020}),
  \bibinfo{pages}{4150--4162}.
\newblock


\bibitem[Coston et~al\mbox{.}(2020b)]%
        {coston2020counterfactual}
\bibfield{author}{\bibinfo{person}{Amanda Coston}, \bibinfo{person}{Alan
  Mishler}, \bibinfo{person}{Edward~H Kennedy}, {and}
  \bibinfo{person}{Alexandra Chouldechova}.} \bibinfo{year}{2020}\natexlab{b}.
\newblock \showarticletitle{Counterfactual risk assessments, evaluation, and
  fairness}. In \bibinfo{booktitle}{\emph{Proceedings of the 2020 conference on
  fairness, accountability, and transparency}}. \bibinfo{pages}{582--593}.
\newblock


\bibitem[Coston et~al\mbox{.}(2022)]%
        {coston2022validity}
\bibfield{author}{\bibinfo{person}{Amanda~Lee Coston}, \bibinfo{person}{Anna
  Kawakami}, \bibinfo{person}{Haiyi Zhu}, \bibinfo{person}{Ken Holstein}, {and}
  \bibinfo{person}{Hoda Heidari}.} \bibinfo{year}{2022}\natexlab{}.
\newblock \showarticletitle{A Validity Perspective on Evaluating the Justified
  Use of Data-driven Decision-making Algorithms}.
\newblock \bibinfo{journal}{\emph{First IEEE Conference on Secure and
  Trustworthy Machine Learning}} (\bibinfo{year}{2022}).
\newblock


\bibitem[De-Arteaga et~al\mbox{.}(2021)]%
        {de2021leveraging}
\bibfield{author}{\bibinfo{person}{Maria De-Arteaga}, \bibinfo{person}{Artur
  Dubrawski}, {and} \bibinfo{person}{Alexandra Chouldechova}.}
  \bibinfo{year}{2021}\natexlab{}.
\newblock \showarticletitle{Leveraging expert consistency to improve
  algorithmic decision support}.
\newblock \bibinfo{journal}{\emph{arXiv preprint arXiv:2101.09648}}
  (\bibinfo{year}{2021}).
\newblock


\bibitem[Denteh and Liebert(2022)]%
        {denteh2022increases}
\bibfield{author}{\bibinfo{person}{Augustine Denteh} {and}
  \bibinfo{person}{Helge Liebert}.} \bibinfo{year}{2022}\natexlab{}.
\newblock \showarticletitle{Who Increases Emergency Department Use? New
  Insights from the Oregon Health Insurance Experiment}.
\newblock \bibinfo{journal}{\emph{arXiv preprint arXiv:2201.07072}}
  (\bibinfo{year}{2022}).
\newblock


\bibitem[Di~Folco et~al\mbox{.}(2022)]%
        {di2022epidemiology}
\bibfield{author}{\bibinfo{person}{C{\'e}cile Di~Folco}, \bibinfo{person}{Ava
  Guez}, \bibinfo{person}{Hugo Peyre}, {and} \bibinfo{person}{Franck Ramus}.}
  \bibinfo{year}{2022}\natexlab{}.
\newblock \showarticletitle{Epidemiology of reading disability: A comparison of
  DSM-5 and ICD-11 criteria}.
\newblock \bibinfo{journal}{\emph{Scientific Studies of Reading}}
  \bibinfo{volume}{26}, \bibinfo{number}{4} (\bibinfo{year}{2022}),
  \bibinfo{pages}{337--355}.
\newblock


\bibitem[D{\'\i}az and van~der Laan(2013)]%
        {diaz2013sensitivity}
\bibfield{author}{\bibinfo{person}{Iv{\'a}n D{\'\i}az} {and}
  \bibinfo{person}{Mark~J van~der Laan}.} \bibinfo{year}{2013}\natexlab{}.
\newblock \showarticletitle{Sensitivity analysis for causal inference under
  unmeasured confounding and measurement error problems}.
\newblock \bibinfo{journal}{\emph{The international journal of biostatistics}}
  \bibinfo{volume}{9}, \bibinfo{number}{2} (\bibinfo{year}{2013}),
  \bibinfo{pages}{149--160}.
\newblock


\bibitem[En{\o}e et~al\mbox{.}(2000)]%
        {enoe2000estimation}
\bibfield{author}{\bibinfo{person}{Claes En{\o}e}, \bibinfo{person}{Marios~P
  Georgiadis}, {and} \bibinfo{person}{Wesley~O Johnson}.}
  \bibinfo{year}{2000}\natexlab{}.
\newblock \showarticletitle{Estimation of sensitivity and specificity of
  diagnostic tests and disease prevalence when the true disease state is
  unknown}.
\newblock \bibinfo{journal}{\emph{Preventive veterinary medicine}}
  \bibinfo{volume}{45}, \bibinfo{number}{1-2} (\bibinfo{year}{2000}),
  \bibinfo{pages}{61--81}.
\newblock


\bibitem[Falagas et~al\mbox{.}(2007)]%
        {falagas2007under}
\bibfield{author}{\bibinfo{person}{ME Falagas}, \bibinfo{person}{KZ Vardakas},
  {and} \bibinfo{person}{PI Vergidis}.} \bibinfo{year}{2007}\natexlab{}.
\newblock \showarticletitle{Under-diagnosis of common chronic diseases:
  prevalence and impact on human health}.
\newblock \bibinfo{journal}{\emph{International journal of clinical practice}}
  \bibinfo{volume}{61}, \bibinfo{number}{9} (\bibinfo{year}{2007}),
  \bibinfo{pages}{1569--1579}.
\newblock


\bibitem[Finkelstein et~al\mbox{.}(2012)]%
        {finkelstein2012oregon}
\bibfield{author}{\bibinfo{person}{Amy Finkelstein}, \bibinfo{person}{Sarah
  Taubman}, \bibinfo{person}{Bill Wright}, \bibinfo{person}{Mira Bernstein},
  \bibinfo{person}{Jonathan Gruber}, \bibinfo{person}{Joseph~P Newhouse},
  \bibinfo{person}{Heidi Allen}, \bibinfo{person}{Katherine Baicker}, {and}
  \bibinfo{person}{Oregon Health~Study Group}.}
  \bibinfo{year}{2012}\natexlab{}.
\newblock \showarticletitle{The Oregon health insurance experiment: evidence
  from the first year}.
\newblock \bibinfo{journal}{\emph{The Quarterly journal of economics}}
  \bibinfo{volume}{127}, \bibinfo{number}{3} (\bibinfo{year}{2012}),
  \bibinfo{pages}{1057--1106}.
\newblock


\bibitem[Finkelstein et~al\mbox{.}(2021)]%
        {finkelstein2021partial}
\bibfield{author}{\bibinfo{person}{Noam Finkelstein}, \bibinfo{person}{Roy
  Adams}, \bibinfo{person}{Suchi Saria}, {and} \bibinfo{person}{Ilya
  Shpitser}.} \bibinfo{year}{2021}\natexlab{}.
\newblock \showarticletitle{Partial identifiability in discrete data with
  measurement error}. In \bibinfo{booktitle}{\emph{Uncertainty in Artificial
  Intelligence}}. PMLR, \bibinfo{pages}{1798--1808}.
\newblock


\bibitem[Fogliato et~al\mbox{.}(2020)]%
        {fogliato2020fairness}
\bibfield{author}{\bibinfo{person}{Riccardo Fogliato},
  \bibinfo{person}{Alexandra Chouldechova}, {and} \bibinfo{person}{Max
  G’Sell}.} \bibinfo{year}{2020}\natexlab{}.
\newblock \showarticletitle{Fairness evaluation in presence of biased noisy
  labels}. In \bibinfo{booktitle}{\emph{International Conference on Artificial
  Intelligence and Statistics}}. PMLR, \bibinfo{pages}{2325--2336}.
\newblock


\bibitem[Fogliato et~al\mbox{.}(2021)]%
        {fogliato2021validity}
\bibfield{author}{\bibinfo{person}{Riccardo Fogliato}, \bibinfo{person}{Alice
  Xiang}, \bibinfo{person}{Zachary Lipton}, \bibinfo{person}{Daniel Nagin},
  {and} \bibinfo{person}{Alexandra Chouldechova}.}
  \bibinfo{year}{2021}\natexlab{}.
\newblock \showarticletitle{On the Validity of Arrest as a Proxy for Offense:
  Race and the Likelihood of Arrest for Violent Crimes}. In
  \bibinfo{booktitle}{\emph{Proceedings of the 2021 AAAI/ACM Conference on AI,
  Ethics, and Society}}. \bibinfo{pages}{100--111}.
\newblock


\bibitem[Gamerman et~al\mbox{.}(2019)]%
        {gamerman2019pragmatic}
\bibfield{author}{\bibinfo{person}{Victoria Gamerman}, \bibinfo{person}{Tianxi
  Cai}, {and} \bibinfo{person}{Amelie Els{\"a}{\ss}er}.}
  \bibinfo{year}{2019}\natexlab{}.
\newblock \showarticletitle{Pragmatic randomized clinical trials: best
  practices and statistical guidance}.
\newblock \bibinfo{journal}{\emph{Health Services and Outcomes Research
  Methodology}}  \bibinfo{volume}{19} (\bibinfo{year}{2019}),
  \bibinfo{pages}{23--35}.
\newblock


\bibitem[Han et~al\mbox{.}(2020)]%
        {han2020survey}
\bibfield{author}{\bibinfo{person}{Bo Han}, \bibinfo{person}{Quanming Yao},
  \bibinfo{person}{Tongliang Liu}, \bibinfo{person}{Gang Niu},
  \bibinfo{person}{Ivor~W Tsang}, \bibinfo{person}{James~T Kwok}, {and}
  \bibinfo{person}{Masashi Sugiyama}.} \bibinfo{year}{2020}\natexlab{}.
\newblock \showarticletitle{A survey of label-noise representation learning:
  Past, present and future}.
\newblock \bibinfo{journal}{\emph{arXiv preprint arXiv:2011.04406}}
  (\bibinfo{year}{2020}).
\newblock


\bibitem[Hill(2011)]%
        {hill2011bayesian}
\bibfield{author}{\bibinfo{person}{Jennifer~L Hill}.}
  \bibinfo{year}{2011}\natexlab{}.
\newblock \showarticletitle{Bayesian nonparametric modeling for causal
  inference}.
\newblock \bibinfo{journal}{\emph{Journal of Computational and Graphical
  Statistics}} \bibinfo{volume}{20}, \bibinfo{number}{1}
  (\bibinfo{year}{2011}), \bibinfo{pages}{217--240}.
\newblock


\bibitem[Hui and Walter(1980)]%
        {hui1980estimating}
\bibfield{author}{\bibinfo{person}{Sui~L Hui} {and} \bibinfo{person}{Steven~D
  Walter}.} \bibinfo{year}{1980}\natexlab{}.
\newblock \showarticletitle{Estimating the error rates of diagnostic tests}.
\newblock \bibinfo{journal}{\emph{Biometrics}} (\bibinfo{year}{1980}),
  \bibinfo{pages}{167--171}.
\newblock


\bibitem[Hur et~al\mbox{.}(2022)]%
        {hur2022using}
\bibfield{author}{\bibinfo{person}{Paul Hur}, \bibinfo{person}{HaeJin Lee},
  \bibinfo{person}{Suma Bhat}, {and} \bibinfo{person}{Nigel Bosch}.}
  \bibinfo{year}{2022}\natexlab{}.
\newblock \showarticletitle{Using Machine Learning Explainability Methods to
  Personalize Interventions for Students.}
\newblock \bibinfo{journal}{\emph{International Educational Data Mining
  Society}} (\bibinfo{year}{2022}).
\newblock


\bibitem[Jacobs and Wallach(2021)]%
        {jacobs2021measurement}
\bibfield{author}{\bibinfo{person}{Abigail~Z Jacobs} {and}
  \bibinfo{person}{Hanna Wallach}.} \bibinfo{year}{2021}\natexlab{}.
\newblock \showarticletitle{Measurement and fairness}. In
  \bibinfo{booktitle}{\emph{Proceedings of the 2021 ACM conference on fairness,
  accountability, and transparency}}. \bibinfo{pages}{375--385}.
\newblock


\bibitem[Johansson et~al\mbox{.}(2020)]%
        {johansson2020generalization}
\bibfield{author}{\bibinfo{person}{Fredrik~D Johansson}, \bibinfo{person}{Uri
  Shalit}, \bibinfo{person}{Nathan Kallus}, {and} \bibinfo{person}{David
  Sontag}.} \bibinfo{year}{2020}\natexlab{}.
\newblock \showarticletitle{Generalization bounds and representation learning
  for estimation of potential outcomes and causal effects}.
\newblock \bibinfo{journal}{\emph{arXiv preprint arXiv:2001.07426}}
  (\bibinfo{year}{2020}).
\newblock


\bibitem[Kallus and Zhou(2018)]%
        {kallus2018residual}
\bibfield{author}{\bibinfo{person}{Nathan Kallus} {and} \bibinfo{person}{Angela
  Zhou}.} \bibinfo{year}{2018}\natexlab{}.
\newblock \showarticletitle{Residual unfairness in fair machine learning from
  prejudiced data}. In \bibinfo{booktitle}{\emph{International Conference on
  Machine Learning}}. PMLR, \bibinfo{pages}{2439--2448}.
\newblock


\bibitem[Kawakami et~al\mbox{.}(2022)]%
        {kawakami2022improving}
\bibfield{author}{\bibinfo{person}{Anna Kawakami}, \bibinfo{person}{Venkatesh
  Sivaraman}, \bibinfo{person}{Hao-Fei Cheng}, \bibinfo{person}{Logan
  Stapleton}, \bibinfo{person}{Yanghuidi Cheng}, \bibinfo{person}{Diana Qing},
  \bibinfo{person}{Adam Perer}, \bibinfo{person}{Zhiwei~Steven Wu},
  \bibinfo{person}{Haiyi Zhu}, {and} \bibinfo{person}{Kenneth Holstein}.}
  \bibinfo{year}{2022}\natexlab{}.
\newblock \showarticletitle{Improving Human-AI Partnerships in Child Welfare:
  Understanding Worker Practices, Challenges, and Desires for Algorithmic
  Decision Support}. In \bibinfo{booktitle}{\emph{CHI Conference on Human
  Factors in Computing Systems}}. \bibinfo{pages}{1--18}.
\newblock


\bibitem[Kennedy(2022)]%
        {kennedy2022semiparametric}
\bibfield{author}{\bibinfo{person}{Edward~H Kennedy}.}
  \bibinfo{year}{2022}\natexlab{}.
\newblock \showarticletitle{Semiparametric doubly robust targeted double
  machine learning: a review}.
\newblock \bibinfo{journal}{\emph{arXiv preprint arXiv:2203.06469}}
  (\bibinfo{year}{2022}).
\newblock


\bibitem[Kleinberg et~al\mbox{.}(2018)]%
        {kleinberg2018human}
\bibfield{author}{\bibinfo{person}{Jon Kleinberg}, \bibinfo{person}{Himabindu
  Lakkaraju}, \bibinfo{person}{Jure Leskovec}, \bibinfo{person}{Jens Ludwig},
  {and} \bibinfo{person}{Sendhil Mullainathan}.}
  \bibinfo{year}{2018}\natexlab{}.
\newblock \showarticletitle{Human decisions and machine predictions}.
\newblock \bibinfo{journal}{\emph{The quarterly journal of economics}}
  \bibinfo{volume}{133}, \bibinfo{number}{1} (\bibinfo{year}{2018}),
  \bibinfo{pages}{237--293}.
\newblock


\bibitem[Kleinberg et~al\mbox{.}(2015)]%
        {kleinberg2015prediction}
\bibfield{author}{\bibinfo{person}{Jon Kleinberg}, \bibinfo{person}{Jens
  Ludwig}, \bibinfo{person}{Sendhil Mullainathan}, {and} \bibinfo{person}{Ziad
  Obermeyer}.} \bibinfo{year}{2015}\natexlab{}.
\newblock \showarticletitle{Prediction policy problems}.
\newblock \bibinfo{journal}{\emph{American Economic Review}}
  \bibinfo{volume}{105}, \bibinfo{number}{5} (\bibinfo{year}{2015}),
  \bibinfo{pages}{491--495}.
\newblock


\bibitem[Kruttschnitt et~al\mbox{.}(2014)]%
        {kruttschnitt2014estimating}
\bibfield{author}{\bibinfo{person}{Candace Kruttschnitt},
  \bibinfo{person}{William~D Kalsbeek}, \bibinfo{person}{Carol~C House},
  {et~al\mbox{.}}} \bibinfo{year}{2014}\natexlab{}.
\newblock \showarticletitle{Estimating the incidence of rape and sexual
  assault}.
\newblock  (\bibinfo{year}{2014}).
\newblock


\bibitem[K{\"u}nzel et~al\mbox{.}(2019)]%
        {kunzel2019metalearners}
\bibfield{author}{\bibinfo{person}{S{\"o}ren~R K{\"u}nzel},
  \bibinfo{person}{Jasjeet~S Sekhon}, \bibinfo{person}{Peter~J Bickel}, {and}
  \bibinfo{person}{Bin Yu}.} \bibinfo{year}{2019}\natexlab{}.
\newblock \showarticletitle{Metalearners for estimating heterogeneous treatment
  effects using machine learning}.
\newblock \bibinfo{journal}{\emph{Proceedings of the national academy of
  sciences}} \bibinfo{volume}{116}, \bibinfo{number}{10}
  (\bibinfo{year}{2019}), \bibinfo{pages}{4156--4165}.
\newblock


\bibitem[Lakkaraju et~al\mbox{.}(2017)]%
        {lakkaraju2017selective}
\bibfield{author}{\bibinfo{person}{Himabindu Lakkaraju}, \bibinfo{person}{Jon
  Kleinberg}, \bibinfo{person}{Jure Leskovec}, \bibinfo{person}{Jens Ludwig},
  {and} \bibinfo{person}{Sendhil Mullainathan}.}
  \bibinfo{year}{2017}\natexlab{}.
\newblock \showarticletitle{The selective labels problem: Evaluating
  algorithmic predictions in the presence of unobservables}. In
  \bibinfo{booktitle}{\emph{Proceedings of the 23rd ACM SIGKDD International
  Conference on Knowledge Discovery and Data Mining}}.
  \bibinfo{pages}{275--284}.
\newblock


\bibitem[LaLonde(1986)]%
        {lalonde1986evaluating}
\bibfield{author}{\bibinfo{person}{Robert~J LaLonde}.}
  \bibinfo{year}{1986}\natexlab{}.
\newblock \showarticletitle{Evaluating the econometric evaluations of training
  programs with experimental data}.
\newblock \bibinfo{journal}{\emph{The American economic review}}
  (\bibinfo{year}{1986}), \bibinfo{pages}{604--620}.
\newblock


\bibitem[Liu and Tao(2015)]%
        {liu2015classification}
\bibfield{author}{\bibinfo{person}{Tongliang Liu} {and}
  \bibinfo{person}{Dacheng Tao}.} \bibinfo{year}{2015}\natexlab{}.
\newblock \showarticletitle{Classification with noisy labels by importance
  reweighting}.
\newblock \bibinfo{journal}{\emph{IEEE Transactions on pattern analysis and
  machine intelligence}} \bibinfo{volume}{38}, \bibinfo{number}{3}
  (\bibinfo{year}{2015}), \bibinfo{pages}{447--461}.
\newblock


\bibitem[Lohr(2019)]%
        {lohr2019measuring}
\bibfield{author}{\bibinfo{person}{Sharon Lohr}.}
  \bibinfo{year}{2019}\natexlab{}.
\newblock \bibinfo{booktitle}{\emph{Measuring crime: Behind the statistics}}.
\newblock \bibinfo{publisher}{Chapman and Hall/CRC}.
\newblock


\bibitem[Lundberg et~al\mbox{.}(2021)]%
        {lundberg2021your}
\bibfield{author}{\bibinfo{person}{Ian Lundberg}, \bibinfo{person}{Rebecca
  Johnson}, {and} \bibinfo{person}{Brandon~M Stewart}.}
  \bibinfo{year}{2021}\natexlab{}.
\newblock \showarticletitle{What is your estimand? Defining the target quantity
  connects statistical evidence to theory}.
\newblock \bibinfo{journal}{\emph{American Sociological Review}}
  \bibinfo{volume}{86}, \bibinfo{number}{3} (\bibinfo{year}{2021}),
  \bibinfo{pages}{532--565}.
\newblock


\bibitem[Menon et~al\mbox{.}(2015)]%
        {menon2015learning}
\bibfield{author}{\bibinfo{person}{Aditya Menon}, \bibinfo{person}{Brendan
  Van~Rooyen}, \bibinfo{person}{Cheng~Soon Ong}, {and} \bibinfo{person}{Bob
  Williamson}.} \bibinfo{year}{2015}\natexlab{}.
\newblock \showarticletitle{Learning from corrupted binary labels via
  class-probability estimation}. In \bibinfo{booktitle}{\emph{International
  conference on machine learning}}. PMLR, \bibinfo{pages}{125--134}.
\newblock


\bibitem[Mullainathan and Obermeyer(2017)]%
        {mullainathan2017does}
\bibfield{author}{\bibinfo{person}{Sendhil Mullainathan} {and}
  \bibinfo{person}{Ziad Obermeyer}.} \bibinfo{year}{2017}\natexlab{}.
\newblock \showarticletitle{Does machine learning automate moral hazard and
  error?}
\newblock \bibinfo{journal}{\emph{American Economic Review}}
  \bibinfo{volume}{107}, \bibinfo{number}{5} (\bibinfo{year}{2017}),
  \bibinfo{pages}{476--480}.
\newblock


\bibitem[Mullainathan and Obermeyer(2021)]%
        {mullainathan2021inequity}
\bibfield{author}{\bibinfo{person}{Sendhil Mullainathan} {and}
  \bibinfo{person}{Ziad Obermeyer}.} \bibinfo{year}{2021}\natexlab{}.
\newblock \showarticletitle{On the inequity of predicting a while hoping for
  B}. In \bibinfo{booktitle}{\emph{AEA Papers and Proceedings}},
  Vol.~\bibinfo{volume}{111}. \bibinfo{pages}{37--42}.
\newblock


\bibitem[Natarajan et~al\mbox{.}(2013)]%
        {natarajan2013learning}
\bibfield{author}{\bibinfo{person}{Nagarajan Natarajan},
  \bibinfo{person}{Inderjit~S Dhillon}, \bibinfo{person}{Pradeep~K Ravikumar},
  {and} \bibinfo{person}{Ambuj Tewari}.} \bibinfo{year}{2013}\natexlab{}.
\newblock \showarticletitle{Learning with noisy labels}.
\newblock \bibinfo{journal}{\emph{Advances in neural information processing
  systems}}  \bibinfo{volume}{26} (\bibinfo{year}{2013}).
\newblock


\bibitem[Nie et~al\mbox{.}(2021)]%
        {nie2021vcnet}
\bibfield{author}{\bibinfo{person}{Lizhen Nie}, \bibinfo{person}{Mao Ye},
  \bibinfo{person}{Qiang Liu}, {and} \bibinfo{person}{Dan Nicolae}.}
  \bibinfo{year}{2021}\natexlab{}.
\newblock \showarticletitle{Vcnet and functional targeted regularization for
  learning causal effects of continuous treatments}.
\newblock \bibinfo{journal}{\emph{arXiv preprint arXiv:2103.07861}}
  (\bibinfo{year}{2021}).
\newblock


\bibitem[Northcutt et~al\mbox{.}(2021)]%
        {northcutt2021confident}
\bibfield{author}{\bibinfo{person}{Curtis Northcutt}, \bibinfo{person}{Lu
  Jiang}, {and} \bibinfo{person}{Isaac Chuang}.}
  \bibinfo{year}{2021}\natexlab{}.
\newblock \showarticletitle{Confident learning: Estimating uncertainty in
  dataset labels}.
\newblock \bibinfo{journal}{\emph{Journal of Artificial Intelligence Research}}
   \bibinfo{volume}{70} (\bibinfo{year}{2021}), \bibinfo{pages}{1373--1411}.
\newblock


\bibitem[Obermeyer et~al\mbox{.}(2019)]%
        {obermeyer2019dissecting}
\bibfield{author}{\bibinfo{person}{Ziad Obermeyer}, \bibinfo{person}{Brian
  Powers}, \bibinfo{person}{Christine Vogeli}, {and} \bibinfo{person}{Sendhil
  Mullainathan}.} \bibinfo{year}{2019}\natexlab{}.
\newblock \showarticletitle{Dissecting racial bias in an algorithm used to
  manage the health of populations}.
\newblock \bibinfo{journal}{\emph{Science}} \bibinfo{volume}{366},
  \bibinfo{number}{6464} (\bibinfo{year}{2019}), \bibinfo{pages}{447--453}.
\newblock


\bibitem[Orso et~al\mbox{.}(2017)]%
        {orso2017epidemiology}
\bibfield{author}{\bibinfo{person}{Francesco Orso}, \bibinfo{person}{Gianna
  Fabbri}, {and} \bibinfo{person}{Aldo~Pietro Maggioni}.}
  \bibinfo{year}{2017}\natexlab{}.
\newblock \showarticletitle{Epidemiology of heart failure}.
\newblock \bibinfo{journal}{\emph{Heart Failure}} (\bibinfo{year}{2017}),
  \bibinfo{pages}{15--33}.
\newblock


\bibitem[Patrini et~al\mbox{.}(2017)]%
        {patrini2017making}
\bibfield{author}{\bibinfo{person}{Giorgio Patrini},
  \bibinfo{person}{Alessandro Rozza}, \bibinfo{person}{Aditya Krishna~Menon},
  \bibinfo{person}{Richard Nock}, {and} \bibinfo{person}{Lizhen Qu}.}
  \bibinfo{year}{2017}\natexlab{}.
\newblock \showarticletitle{Making deep neural networks robust to label noise:
  A loss correction approach}. In \bibinfo{booktitle}{\emph{Proceedings of the
  IEEE conference on computer vision and pattern recognition}}.
  \bibinfo{pages}{1944--1952}.
\newblock


\bibitem[Pearl(2009)]%
        {pearl2009causal}
\bibfield{author}{\bibinfo{person}{Judea Pearl}.}
  \bibinfo{year}{2009}\natexlab{}.
\newblock \showarticletitle{Causal inference in statistics: An overview}.
\newblock \bibinfo{journal}{\emph{Statistics surveys}}  \bibinfo{volume}{3}
  (\bibinfo{year}{2009}), \bibinfo{pages}{96--146}.
\newblock


\bibitem[Perdomo et~al\mbox{.}(2020)]%
        {perdomo2020performative}
\bibfield{author}{\bibinfo{person}{Juan Perdomo}, \bibinfo{person}{Tijana
  Zrnic}, \bibinfo{person}{Celestine Mendler-D{\"u}nner}, {and}
  \bibinfo{person}{Moritz Hardt}.} \bibinfo{year}{2020}\natexlab{}.
\newblock \showarticletitle{Performative prediction}. In
  \bibinfo{booktitle}{\emph{International Conference on Machine Learning}}.
  PMLR, \bibinfo{pages}{7599--7609}.
\newblock


\bibitem[Raji et~al\mbox{.}(2022)]%
        {raji2022fallacy}
\bibfield{author}{\bibinfo{person}{Inioluwa~Deborah Raji},
  \bibinfo{person}{I~Elizabeth Kumar}, \bibinfo{person}{Aaron Horowitz}, {and}
  \bibinfo{person}{Andrew Selbst}.} \bibinfo{year}{2022}\natexlab{}.
\newblock \showarticletitle{The fallacy of AI functionality}. In
  \bibinfo{booktitle}{\emph{2022 ACM Conference on Fairness, Accountability,
  and Transparency}}. \bibinfo{pages}{959--972}.
\newblock


\bibitem[Rambachan et~al\mbox{.}(2022)]%
        {rambachan2022counterfactual}
\bibfield{author}{\bibinfo{person}{Ashesh Rambachan}, \bibinfo{person}{Amanda
  Coston}, {and} \bibinfo{person}{Edward Kennedy}.}
  \bibinfo{year}{2022}\natexlab{}.
\newblock \showarticletitle{Counterfactual Risk Assessments under Unmeasured
  Confounding}.
\newblock \bibinfo{journal}{\emph{arXiv preprint arXiv:2212.09844}}
  (\bibinfo{year}{2022}).
\newblock


\bibitem[Reeve et~al\mbox{.}(2019)]%
        {reeve2019classification}
\bibfield{author}{\bibinfo{person}{Henry Reeve} {et~al\mbox{.}}}
  \bibinfo{year}{2019}\natexlab{}.
\newblock \showarticletitle{Classification with unknown class-conditional label
  noise on non-compact feature spaces}. In \bibinfo{booktitle}{\emph{Conference
  on Learning Theory}}. PMLR, \bibinfo{pages}{2624--2651}.
\newblock


\bibitem[Roberts(1985)]%
        {roberts1985measurement}
\bibfield{author}{\bibinfo{person}{Fred~S Roberts}.}
  \bibinfo{year}{1985}\natexlab{}.
\newblock \showarticletitle{Measurement theory}.
\newblock  (\bibinfo{year}{1985}).
\newblock


\bibitem[Robins(1986)]%
        {robins1986new}
\bibfield{author}{\bibinfo{person}{James Robins}.}
  \bibinfo{year}{1986}\natexlab{}.
\newblock \showarticletitle{A new approach to causal inference in mortality
  studies with a sustained exposure period—application to control of the
  healthy worker survivor effect}.
\newblock \bibinfo{journal}{\emph{Mathematical modelling}} \bibinfo{volume}{7},
  \bibinfo{number}{9-12} (\bibinfo{year}{1986}), \bibinfo{pages}{1393--1512}.
\newblock


\bibitem[Rosenbaum and Rubin(1983)]%
        {rosenbaum1983central}
\bibfield{author}{\bibinfo{person}{Paul~R Rosenbaum} {and}
  \bibinfo{person}{Donald~B Rubin}.} \bibinfo{year}{1983}\natexlab{}.
\newblock \showarticletitle{The central role of the propensity score in
  observational studies for causal effects}.
\newblock \bibinfo{journal}{\emph{Biometrika}} \bibinfo{volume}{70},
  \bibinfo{number}{1} (\bibinfo{year}{1983}), \bibinfo{pages}{41--55}.
\newblock


\bibitem[Rubin(1974)]%
        {rubin1974estimating}
\bibfield{author}{\bibinfo{person}{Donald~B Rubin}.}
  \bibinfo{year}{1974}\natexlab{}.
\newblock \showarticletitle{Estimating causal effects of treatments in
  randomized and nonrandomized studies.}
\newblock \bibinfo{journal}{\emph{Journal of educational Psychology}}
  \bibinfo{volume}{66}, \bibinfo{number}{5} (\bibinfo{year}{1974}),
  \bibinfo{pages}{688}.
\newblock


\bibitem[Rubin(2005)]%
        {rubin2005causal}
\bibfield{author}{\bibinfo{person}{Donald~B Rubin}.}
  \bibinfo{year}{2005}\natexlab{}.
\newblock \showarticletitle{Causal inference using potential outcomes: Design,
  modeling, decisions}.
\newblock \bibinfo{journal}{\emph{J. Amer. Statist. Assoc.}}
  \bibinfo{volume}{100}, \bibinfo{number}{469} (\bibinfo{year}{2005}),
  \bibinfo{pages}{322--331}.
\newblock


\bibitem[Schouwenburg(2004)]%
        {schouwenburg2004procrastination}
\bibfield{author}{\bibinfo{person}{Henri~C Schouwenburg}.}
  \bibinfo{year}{2004}\natexlab{}.
\newblock \showarticletitle{Procrastination in Academic Settings: General
  Introduction.}
\newblock  (\bibinfo{year}{2004}).
\newblock


\bibitem[Scott(2015)]%
        {scott2015rate}
\bibfield{author}{\bibinfo{person}{Clayton Scott}.}
  \bibinfo{year}{2015}\natexlab{}.
\newblock \showarticletitle{A rate of convergence for mixture proportion
  estimation, with application to learning from noisy labels}. In
  \bibinfo{booktitle}{\emph{Artificial Intelligence and Statistics}}. PMLR,
  \bibinfo{pages}{838--846}.
\newblock


\bibitem[Scott et~al\mbox{.}(2013)]%
        {scott2013classification}
\bibfield{author}{\bibinfo{person}{Clayton Scott}, \bibinfo{person}{Gilles
  Blanchard}, {and} \bibinfo{person}{Gregory Handy}.}
  \bibinfo{year}{2013}\natexlab{}.
\newblock \showarticletitle{Classification with asymmetric label noise:
  Consistency and maximal denoising}. In \bibinfo{booktitle}{\emph{Conference
  on learning theory}}. PMLR, \bibinfo{pages}{489--511}.
\newblock


\bibitem[Shalit et~al\mbox{.}(2017)]%
        {shalit2017estimating}
\bibfield{author}{\bibinfo{person}{Uri Shalit}, \bibinfo{person}{Fredrik~D
  Johansson}, {and} \bibinfo{person}{David Sontag}.}
  \bibinfo{year}{2017}\natexlab{}.
\newblock \showarticletitle{Estimating individual treatment effect:
  generalization bounds and algorithms}. In
  \bibinfo{booktitle}{\emph{International Conference on Machine Learning}}.
  PMLR, \bibinfo{pages}{3076--3085}.
\newblock


\bibitem[Shi et~al\mbox{.}(2019)]%
        {shi2019adapting}
\bibfield{author}{\bibinfo{person}{Claudia Shi}, \bibinfo{person}{David Blei},
  {and} \bibinfo{person}{Victor Veitch}.} \bibinfo{year}{2019}\natexlab{}.
\newblock \showarticletitle{Adapting neural networks for the estimation of
  treatment effects}.
\newblock \bibinfo{journal}{\emph{Advances in neural information processing
  systems}}  \bibinfo{volume}{32} (\bibinfo{year}{2019}).
\newblock


\bibitem[Shimodaira(2000)]%
        {shimodaira2000improving}
\bibfield{author}{\bibinfo{person}{Hidetoshi Shimodaira}.}
  \bibinfo{year}{2000}\natexlab{}.
\newblock \showarticletitle{Improving predictive inference under covariate
  shift by weighting the log-likelihood function}.
\newblock \bibinfo{journal}{\emph{Journal of statistical planning and
  inference}} \bibinfo{volume}{90}, \bibinfo{number}{2} (\bibinfo{year}{2000}),
  \bibinfo{pages}{227--244}.
\newblock


\bibitem[Shrout and Lane(2012)]%
        {shrout2012psychometrics}
\bibfield{author}{\bibinfo{person}{Patrick~E Shrout} {and}
  \bibinfo{person}{Sean~P Lane}.} \bibinfo{year}{2012}\natexlab{}.
\newblock \showarticletitle{Psychometrics.}
\newblock  (\bibinfo{year}{2012}).
\newblock


\bibitem[Shu and Yi(2019)]%
        {shu2019causal}
\bibfield{author}{\bibinfo{person}{Di Shu} {and} \bibinfo{person}{Grace~Y Yi}.}
  \bibinfo{year}{2019}\natexlab{}.
\newblock \showarticletitle{Causal inference with measurement error in
  outcomes: Bias analysis and estimation methods}.
\newblock \bibinfo{journal}{\emph{Statistical methods in medical research}}
  \bibinfo{volume}{28}, \bibinfo{number}{7} (\bibinfo{year}{2019}),
  \bibinfo{pages}{2049--2068}.
\newblock


\bibitem[Smith and Todd(2005)]%
        {smith2005does}
\bibfield{author}{\bibinfo{person}{Jeffrey~A Smith} {and}
  \bibinfo{person}{Petra~E Todd}.} \bibinfo{year}{2005}\natexlab{}.
\newblock \showarticletitle{Does matching overcome LaLonde's critique of
  nonexperimental estimators?}
\newblock \bibinfo{journal}{\emph{Journal of econometrics}}
  \bibinfo{volume}{125}, \bibinfo{number}{1-2} (\bibinfo{year}{2005}),
  \bibinfo{pages}{305--353}.
\newblock


\bibitem[Turque(2012)]%
        {turque2012creative}
\bibfield{author}{\bibinfo{person}{Bill Turque}.}
  \bibinfo{year}{2012}\natexlab{}.
\newblock \showarticletitle{Creative... motivating’and fired}.
\newblock \bibinfo{journal}{\emph{The Washington Post}}  \bibinfo{volume}{6}
  (\bibinfo{year}{2012}).
\newblock


\bibitem[Van~Rooyen et~al\mbox{.}(2015a)]%
        {van2015machine}
\bibfield{author}{\bibinfo{person}{Brendan Van~Rooyen} {et~al\mbox{.}}}
  \bibinfo{year}{2015}\natexlab{a}.
\newblock \showarticletitle{Machine learning via transitions}.
\newblock  (\bibinfo{year}{2015}).
\newblock


\bibitem[Van~Rooyen et~al\mbox{.}(2015b)]%
        {van2015learning}
\bibfield{author}{\bibinfo{person}{Brendan Van~Rooyen}, \bibinfo{person}{Aditya
  Menon}, {and} \bibinfo{person}{Robert~C Williamson}.}
  \bibinfo{year}{2015}\natexlab{b}.
\newblock \showarticletitle{Learning with symmetric label noise: The importance
  of being unhinged}.
\newblock \bibinfo{journal}{\emph{Advances in neural information processing
  systems}}  \bibinfo{volume}{28} (\bibinfo{year}{2015}).
\newblock


\bibitem[Walter and Irwig(1988)]%
        {walter1988estimation}
\bibfield{author}{\bibinfo{person}{Steven~D Walter} {and}
  \bibinfo{person}{Les~M Irwig}.} \bibinfo{year}{1988}\natexlab{}.
\newblock \showarticletitle{Estimation of test error rates, disease prevalence
  and relative risk from misclassified data: a review}.
\newblock \bibinfo{journal}{\emph{Journal of clinical epidemiology}}
  \bibinfo{volume}{41}, \bibinfo{number}{9} (\bibinfo{year}{1988}),
  \bibinfo{pages}{923--937}.
\newblock


\bibitem[Wang et~al\mbox{.}(2022)]%
        {wang2022against}
\bibfield{author}{\bibinfo{person}{Angelina Wang}, \bibinfo{person}{Sayash
  Kapoor}, \bibinfo{person}{Solon Barocas}, {and} \bibinfo{person}{Arvind
  Narayanan}.} \bibinfo{year}{2022}\natexlab{}.
\newblock \showarticletitle{Against Predictive Optimization: On the Legitimacy
  of Decision-Making Algorithms that Optimize Predictive Accuracy}.
\newblock \bibinfo{journal}{\emph{Available at SSRN}} (\bibinfo{year}{2022}).
\newblock


\bibitem[Wang et~al\mbox{.}(2021)]%
        {wang2021fair}
\bibfield{author}{\bibinfo{person}{Jialu Wang}, \bibinfo{person}{Yang Liu},
  {and} \bibinfo{person}{Caleb Levy}.} \bibinfo{year}{2021}\natexlab{}.
\newblock \showarticletitle{Fair classification with group-dependent label
  noise}. In \bibinfo{booktitle}{\emph{Proceedings of the 2021 ACM conference
  on fairness, accountability, and transparency}}. \bibinfo{pages}{526--536}.
\newblock


\bibitem[Xia et~al\mbox{.}(2020)]%
        {xia2020part}
\bibfield{author}{\bibinfo{person}{Xiaobo Xia}, \bibinfo{person}{Tongliang
  Liu}, \bibinfo{person}{Bo Han}, \bibinfo{person}{Nannan Wang},
  \bibinfo{person}{Mingming Gong}, \bibinfo{person}{Haifeng Liu},
  \bibinfo{person}{Gang Niu}, \bibinfo{person}{Dacheng Tao}, {and}
  \bibinfo{person}{Masashi Sugiyama}.} \bibinfo{year}{2020}\natexlab{}.
\newblock \showarticletitle{Part-dependent label noise: Towards
  instance-dependent label noise}.
\newblock \bibinfo{journal}{\emph{Advances in Neural Information Processing
  Systems}}  \bibinfo{volume}{33} (\bibinfo{year}{2020}),
  \bibinfo{pages}{7597--7610}.
\newblock


\bibitem[Xia et~al\mbox{.}(2019)]%
        {xia2019anchor}
\bibfield{author}{\bibinfo{person}{Xiaobo Xia}, \bibinfo{person}{Tongliang
  Liu}, \bibinfo{person}{Nannan Wang}, \bibinfo{person}{Bo Han},
  \bibinfo{person}{Chen Gong}, \bibinfo{person}{Gang Niu}, {and}
  \bibinfo{person}{Masashi Sugiyama}.} \bibinfo{year}{2019}\natexlab{}.
\newblock \showarticletitle{Are anchor points really indispensable in
  label-noise learning?}
\newblock \bibinfo{journal}{\emph{Advances in Neural Information Processing
  Systems}}  \bibinfo{volume}{32} (\bibinfo{year}{2019}).
\newblock


\bibitem[Zwaan and Singh(2015)]%
        {zwaan2015challenges}
\bibfield{author}{\bibinfo{person}{Laura Zwaan} {and} \bibinfo{person}{Hardeep
  Singh}.} \bibinfo{year}{2015}\natexlab{}.
\newblock \showarticletitle{The challenges in defining and measuring diagnostic
  error}.
\newblock \bibinfo{journal}{\emph{Diagnosis}} \bibinfo{volume}{2},
  \bibinfo{number}{2} (\bibinfo{year}{2015}), \bibinfo{pages}{97--103}.
\newblock


\end{thebibliography}

\clearpage
\appendix
\section{Appendix}\label{sec:appendix}

This appendix contains the following subsections: 

\begin{itemize}
    \item  \ref{sec:appendix_obermeyer} provides a discussion of our re-analysis of audit data released by \citet{obermeyer2019dissecting}. 
    \item  \ref{sec:appendix_proof} contains omitted proofs for theorems introduced in $\S$ \ref{sec:methodology}. 
    \item \ref{sec:appendix_alg_details} contains omitted algorithm pseudocode.
    \item \ref{sec:appendix_experiments} contains additional details \lgedit{and results} for experiments reported in Section \ref{sec:experiments}. 
\end{itemize}

\begin{table}
\centering
\begin{tabular}{lrrr}
\hline
Sample      &   FPR &   FNR &  N \\
\hline
Full population &  0.37 &  0.38 & 48,784 \\
Unenrolled      &  0.37 &  0.39 & 48,332 \\ 
Enrolled        & \textcolor{darkred}{0.64} &  \textcolor{darkred}{0.13} &   452 \\
\hline
\end{tabular}
\caption{Treatment-conditional OME parameters computed using synthetic data released by \citet{obermeyer2019dissecting}.}\label{table:obsermeyer_rates}
\end{table}

\begin{table}
\centering
\begin{tabular}{lrrr}
\hline
Sample      &   FPR &   FNR &     N \\
\hline
Full population &  0.36 &  0.39 & 48,784 \\
Unenrolled      &  0.36 &  0.39 & 48,360 \\
Enrolled        &  \textcolor{darkred}{0.65} &  \textcolor{darkred}{0.14} &   424 \\
\hline
\end{tabular}
\caption{Treatment-conditional OME parameters computed after re-applying synthpop on released synthetic data.}\label{table:synthpop_rates}
\end{table}

\subsection{Re-analysis of data published by \citet{obermeyer2019dissecting}}\label{sec:appendix_obermeyer}

\citet{obermeyer2019dissecting} release publicly available synthetic dataset corresponding to their audit of a clinical risk assessment.\footnote{https://gitlab.com/labsysmed/dissecting-bias} Synthetic data was generated via the R package synthpop, which preserves moments and covariances of the original dataset. The synthetic data release is sufficient to replicate the main analyses reported over the raw (unmodified dataset) reported in \citep{obermeyer2019dissecting}. This makes it likely that our analysis closely preserves the true statistics reported on raw data, as our only analysis step involves thresholding raw scores and computing conditional probabilities.

We probe the implications of naively estimating population OME parameters by reanalyzing public synthetic data published as part of the audit study. Our analysis estimates proxy error parameters by binarizing continuous cost ($Y$) and chronic active condition ($Y^*$) outcomes at the 55th risk percentile: the threshold used in practice to drive enrollment recommendations. While this choice of target outcome is itself imperfect \cite{falagas2007under}, we use chronic active conditions as a reference outcome to match the original comparison conducted by \citet{obermeyer2019dissecting}.

Our analysis (Table \ref{table:obsermeyer_rates}) finds that the false positive and false negative rates of the cost of care proxy varies substantially across program enrollment status. In particular, the false negative rate is \textcolor{darkred}{65.8\%} lower among patients enrolled in the program as compared to the full population, while the FPR is \textcolor{darkred}{72.9\%} higher. This difference is consistent with closer medical supervision: under enrollment, patients may incur greater costs due to expanded care, even after controlling for the number of underlying active chronic conditions. In contrast, OME parameters among the unenrolled resemble the population average because the vast majority of patients ($\approx 99\%$) are turned away from the program. We verify that this finding is not an artifact of synthetic data generation by re-applying synthpop on data provided by \citep{obermeyer2019dissecting} and re-computing error parameters via the same procedure described above (Table \ref{table:synthpop_rates}). While we observe minor variations in error parameters after re-applying synthpop, the large difference in error rates across the full population and enrollment conditions persists. 

Triangulating the downstream impacts of this error parameter discrepancy is challenging. To preserve privacy, the research team did not release covariates needed to re-train an algorithm. Prior program enrollment decisions were also non-randomized, meaning that differences in error parameters could be attributed to unmeasured confounders. Nevertheless, the difference in error parameters across enrolled and unenrolled carries serious implications for the diagnosis and mitigation of outcome measurement error.\footnote{\lgedit{\citet{obermeyer2019dissecting} report robustness checks examining whether differential program effects by race could impact their study of label bias. The authors found no such differential effects by race. As a result, their main analyses are not likely to be impacted by the findings of our re-analysis. Nevertheless, the model reliability challenges we study in this work could impact \textit{all individuals in the study population}, if unaddressed.}}

\newpage
\subsection{Omitted results and proofs}\label{sec:appendix_proof}

We begin by providing a roof of Theorem \ref{thm:unbiased_risk}. This proof follows from unbiased risk minimization results from the label noise \citep{natarajan2013learning, patrini2017making, chou2020unbiased, van2015machine} and counterfactual prediction \citep{johansson2020generalization} literature. 

\begin{proof}

We will show that $R_{t, \tilde{\ell}}^w(f_t) = R_{\tilde{\ell}}(f_t) = R_{\ell}^*(f_t)$, $\forall t = \{0, 1 \}$. We begin by showing the first equality. We have that
\begin{align*}
R_{t,\tilde{\ell}}^{w}\left(f_t\right)&:=\mathbb{E}_{p}\left[w(X) \tilde{\ell}(f_t(X), Y) \mid T=t\right] \\ &=\mathbb{E}_{p^*}\left[w(X) \tilde{\ell}(f_t(X), Y_t) \mid T=t\right] \\ &= \mathbb{E}_{p^*}\left[w(X) \tilde{\ell}(f_t(X), Y_t) \right]\nonumber    
\end{align*}

where the first equality holds by consistency (\ref{assumption:consistency}) and the second by ignorability (\ref{assumption:ignorability}). As a result, we can express both equalities over potential outcomes $Y_t, Y^*_t \sim p^*$. Next, let $p_t(X) \coloneqq p(X|T=t)$ and let $\tilde{\ell}_{f_t}(x) \coloneqq \mathbb{E}_{Y_t} [ \tilde{\ell}(f_t(x), Y_t)\mid X = x]$ be the \textit{expected pointwise surrogate loss} of $f_t$ evaluated at $x$. Then by Lemma 2 of \citep{johansson2020generalization}, we have

\begin{align*}
R^w_{t, \tilde{\ell}}(f_t) &= \int_{x\in X} w(x) \tilde{\ell}_{f_t}(x) p_t(x) dx \\ &= \int_{x\in X} \frac{p_t(x)}{p(x)} w(x) \tilde{\ell}_{f_t}(x) p(x) dx \\ &= R_{\tilde{\ell}}(f_t)\nonumber
\end{align*}

for $w(x) = p(x)/p_t(x)$. The second equality assumes positivity (\ref{assumption:positivity}) and ignorability (\ref{assumption:ignorability}). Applying Bayes' 
 to $p(x)/p_t(x)$ and rearranging

\begin{equation}
    w(x) = \frac{p(x)}{p(X=x|T=t)} = \frac{p(T=t)}{ p(T=t|X=x)} = \frac{p(T=t)}{(2 t-1) \cdot \pi(x)+1-t}\nonumber 
\end{equation}

which is the weighting function in (\ref{eq:re-weighting}). Next, we show that $R_{\tilde{\ell}}(f_t) = R_{\ell}^*(f_t)$, which follows from Lemma 1 of \cite{natarajan2013learning}. Given

$$
\boldsymbol{\eta_t}(x) =  \left(\begin{array}{cc}
1-\eta_t(x) \\
\eta_t(x)
\end{array}\right),\;\;
\boldsymbol{T}  = \left(\begin{array}{cc}
1-\alpha_t & \alpha_t \\
\beta_t & 1-\beta_t
\end{array}\right)
$$

we can express (\ref{eq:tce_model}) as $\boldsymbol{\eta_t}(x) = \boldsymbol{T} \boldsymbol{\eta^*_t}(x)$ by assumption \ref{assumption:error_model}. This error model is invertable via $\boldsymbol{\eta^*_t}(x) = \boldsymbol{T}^{-1} \boldsymbol{\eta_t}(x)$ for

$$
\boldsymbol{T}^{-1} = \frac{1}{1-\alpha_t-\beta_t}\left(\begin{array}{cc}
1-\beta_t & -\alpha_t \\
-\beta_t & 1-\alpha_t
\end{array}\right).
$$

Let $\boldsymbol{\ell}(f_t(x)) = (\ell(f_t(x), 0), \ell(f_t(x), 1))^\top$ be a vectorized loss corresponding to labels $\boldsymbol{e} \in \{0, 1\}^2$. Then we have that
\begin{align}
R^*_{\ell}(f_t) &=  \mathbb{E}_X \mathbb{E}_{Y^*_t \sim \; \boldsymbol{\eta}_t^*(X)} [\ell(f_t(X), Y^*_t)] = \mathbb{E}_X [ \boldsymbol{\eta}_t^*(X)^\top \boldsymbol{\ell}(f_t(X))]\nonumber\\
&= \mathbb{E}_X [ \boldsymbol{\eta}_t(X)^\top (\mathbf{T}^{-1})\boldsymbol{\ell}(f_t(X))] = \mathbb{E}_X [ \mathbf{e}^\top (\mathbf{T}^{-1})\boldsymbol{\ell}(f_t(X))]\label{eq:app_invs_model} \nonumber\\
&= \mathbb{E}_{X, Y_t}[ \tilde{\ell}(f_t(X), Y_t)] = R_{\tilde{\ell}}(f_t)\nonumber
\end{align}
Therefore, $R_{\tilde{\ell}}(f_t) = R_{\ell}^*(f_t)$ for a surrogate loss constructed via $\boldsymbol{\tilde{\ell}} = (\mathbf{T}^{-1})\boldsymbol{\ell}(f_t(X))$. Multiplying $\boldsymbol{\ell}(f_t(X))$ by $\mathbf{T}^{-1}$ and rearranging yields (\ref{eq:surrogate_loss}).
\end{proof}

Next, we prove Theorem \ref{theorem:anchor_identification} showing that error parameters are identifiable under combinations of assumptions stated in Table \ref{fig:identification_table}.

\begin{proof}

To begin, observe that the error model (\ref{eq:tce_model}) expresses the conditional proxy class probability $\eta_t$ as a linear function of $\eta^*_t$ with two unknowns. Therefore, given knowledge of the target class probability $c^*_{t,i} = \eta^*_t(x_i)$ and proxy class probability $c_{t, i} = \eta_t(x_i)$ at two distinct points $(c^*_{t,i}, c_{t, i})$ and $(c^*_{t,j}, c_{t,j})$, we can set up a linear equation 

\begin{equation}\begin{aligned}\label{eq:identification_system}
c_{t, i} &= (1-\beta_t) \cdot c_{t, i}^* + \alpha_t \cdot (1 -  c_{t, i}^*)\\
c_{t,j} &= (1-\beta_t) \cdot c_{t,j}^* + \alpha_t \cdot (1 - c_{t,j}^*)\\
\end{aligned}\end{equation}
and solve for error parameters
\begin{align}\label{eq:closed_form}
\alpha_t &= \frac{c_{t, i}^* \cdot c_{t,j} - c_{t, i} \cdot c_{t,j}^*}{c^*_{t,i} - c^*_{t,j}} \\ \beta_t &= \frac{c_{t, i} \cdot c_{t,j}^*   - c_{t, i}+c_{t, i}^* - c_{t,j}^*+c_{t,j} - c_{t, i}^* \cdot c_{t,j}}{c_{t, i}^*-c_{t,j}^*}
\end{align}

provided that $c^*_{t,i} \neq c^*_{t,j}$. Identification of the specific cases in Table \ref{fig:identification_table} follows from application of (\ref{eq:closed_form}). When $\alpha_t$ and $\beta_t$ are both known, identification is not required. When one of $\beta_t$ ($\alpha_t$) is known, the corresponding $\alpha_t$ ($\beta_t$) can be given by

\begin{equation}\label{eq:single_param}
\alpha_t = \frac{c_{t, i} - (1-\beta_t) \cdot c_{t, i}^*}{(1-c_{t, i}^*)}, \;\; \beta_t = \frac{c^*_{t,i} - c_{t,i} + \alpha_t \cdot (1-c^*_{t,i})}{c^*_{t,i}}
\end{equation}

Therefore, only one anchor assumption $(c^*_{t,i}, c_{t,i})$ is required given knowledge of $\alpha_t$ or $\beta_t$. However, by (\ref{eq:single_param}), note that $c^*_{t,i} \neq 1$ is required for identification of $\alpha_t$ and $c^*_{t,j} \neq 0$ is required for identification of $\beta_t$. This rules out combinations denoted by (\xmark) in Table \ref{fig:identification_table}. Error parameters can be derived directly from (\ref{eq:closed_form}) if $\alpha_t$ and $\beta_t$ are both unknown so long as $c_{t,i}^* \neq  c_{t,j}^*$. The specific values of $(c_{t,i}, c_{t,i}^*)$ corresponding to min, max, and base rate anchors can be computed via
\begin{align*}
     c_{t,i}^* &= \inf_{x_i \in X} { \{ \eta^*_t(x_i) \} },\quad  c_{t,i} = \inf_{x_i \in \mathcal{X}} \{ \eta_t(x_i)\} \quad \text{(Min anchor)}\\
     c_{t,i}^* &= \sup_{x_i \in X} { \{ \eta^*_t(x_i) \} },\quad  c_{t,i} = \sup_{x_i \in \mathcal{X}} \{ \eta_t(x_i)\} \quad \text{(Max anchor)}\\
     c_{t,i}^* &= \mathbb{E}[\eta^*_t(X)],\quad\quad\;\; c_{t,i} = \mathbb{E}[\eta_t(X)] \quad\quad \text{(Base rate anchor)}
\end{align*}

Above, the min anchor holds because $\eta_t$ is a strictly monotone increasing transform of $\eta^*_t$ by \ref{eq:tce_model} such that  $c_{t,i} = \arginf_{x_i\in X}\{\eta_t(x)\} = \arginf_{x_i\in X}\{\eta^*_t(x)\}$. The max anchor holds by the same argument. The base rate anchor holds because $\mathbb{E}_X[\eta_t(x)] = \mathbb{E}_X[\eta^*_t(x) \cdot (1- \beta_t - \alpha_t) + \alpha_t]$.

Finally, observe that $\eta_t(x)$ is defined over potential outcomes $Y_t \sim p^*$ rather than observational proxies $Y \sim p$. Identification of $\eta_t$ from observational data follows from
\begin{equation}\label{eq:app_identifiability}
\eta_t(x) := p(Y_t=1|X=x) = p(Y=1|X=x, T=t)
\end{equation}
where the equality holds by ignorability (\ref{assumption:ignorability}) and consistency (\ref{assumption:consistency}). By positivity (\ref{assumption:positivity}), we have that the support of $\eta_t(x)$ is defined $\forall x \in X$, which guarantees that the min and max anchor will be defined. 

\end{proof}

\newpage
\subsection{Algorithms}\label{sec:appendix_alg_details}

The RW-SL and CCPE algorithms presented in $\S$ \ref{sec:methodology} partition training data into disjoint folds to learn $\hat{\alpha}_t$, $\hat{\beta}_t$, $\hat{\pi}$, and minimize the re-weighted surrogate risk. We also provide a version of these algorithms with cross-fitting to improve data efficiency. Cross-fitting is useful when using limited data to fit multiple nuisance functions and improves data efficiency while limiting over-fitting \citep{kennedy2022semiparametric}.

\begin{algorithm}
    \SetKwInput{Input}{Input}
    \SetKwInput{Output}{Output}
    \caption{Re-weighted risk minimization with surrogate loss (cross fitting)}\label{alg:reweight_surrogate}
    \Input{Data $\mathcal{W} = \{(X_i, T_i, Y_i)\}_{i=1}^n \sim p$}
    \Output{Learned estimator $\hat{\eta}^*_t(x)$ }
    Partition $\mathcal{W}$ into $\mathcal{W}_1$, $\mathcal{W}_2$, $\mathcal{W}_3$ \\

    \For{$(m, n, p) \in \{ (1,2,3), (3,1,2), (2,3,1) \}$ }{   
       On $\mathcal{W}_m$, estimate parameters $\hat{\alpha}^m_t, \hat{\beta}^m_t \leftarrow \text{CCPE}(\mathcal{W}_m)$\\
       On $\mathcal{W}_n$, learn $\hat{\pi}_n(x)$ by regressing $T \sim X$ \\
       On $\mathcal{W}_p$, use $\hat{\pi}_n(x), \hat{\alpha}^m_t, \hat{\beta}^m_t$ to solve $\hat{\eta}^*_{t,p}(x)  \leftarrow \argmin_{f_t \in \mathcal{H}} \hat{R}^{\hat{w}}_{t, \tilde{\ell}}(f_t)$ \\
    }
    Return combined predictions $\hat{\eta}^*_t(x) = \frac{1}{3} \sum_{p=1}^3 \hat{\eta}^*_{t,p}(x)$ \\
\end{algorithm}\label{alg:rw_sl_crossfit}
\begin{algorithm}
    \SetKwInput{Input}{Input}
    \SetKwInput{Output}{Output}
    \caption{Conditional class probability estimation (cross fitting)}\label{alg:ccpe_crossfit}
    \Input{Data $\mathcal{V} \sim p$}
    \Output{Parameter estimates $\hat{\alpha}_t$, $\hat{\beta}_t$}
    Partition $\mathcal{V}$ into $\mathcal{V}_1$, $\mathcal{V}_2$ \\
    \For{$(m, n) \in \{ (1,2), (2, 1) \}$ }{   
       On $\mathcal{V}_m$, learn $\hat{\eta}_t^m(x)$ by regressing $Y \sim X \; | \; T = t$\\
       On $\mathcal{V}_n$, estimate error parameters: $\hat{\alpha}_t^n = \underset{{x \in X}}{\min} \{ \hat{\eta}_t^m(x) \},\;\; \hat{\beta}_t^n = 1 - \underset{{x \in X}}{\max}  \{ \hat{\eta}_t^m(x) \} $
    }
    Return averaged parameters $\hat{\alpha}_t = \frac{1}{2} \sum_{n=1}^2 \hat{\alpha}_t^n$, $\hat{\beta}_t = 
 \frac{1}{2} \sum_{n=1}^2 \hat{\beta}_t^n$
\end{algorithm}

\clearpage
\subsection{Additional experimental details and results}\label{sec:appendix_experiments}

\subsubsection{Setup details}

 In our synthetic evaluation, we sample from target class probability functions $\eta_0^*(x) = .4+.4\cos(9x+5.5), \forall x \in [-1, -.237]$;  $.5+.3\sin(8x+.9)+.15\sin(10x+.2)+.05\sin(30x+.2), \forall x \in (-.237, 1]$ and $\eta_1^*(x) = .5 - .5 \sin(2.9x + .1)$ and sample treatments from the linear function $\pi(x) = .35 x + .5$ (Figure \ref{fig:syn_setup}).

We train all models with a binary-cross entropy loss. We use the same 4-layer MLP implemented via PyTorch with hidden layer sizes $(40, 30, 10)$ for all models discussed in $\S$ \ref{subsec:baselines}. Where relevant, we also fit $\pi(x)$ and $\eta_t(x)$ (used in CCPE) via the same architecture. We train all models for 10 each epochs each at learning rate $\eta=.5e^{-3}$. Hyperparameters were selected via a hyperparameter sweep optimizing accuracy on $Y^*_0$ with respect to the TPO model.

In our semi-synthetic experiments, we run all models in the synthetic experiment without sample splitting and cross-fitting. \lgedit{While cross-fitting improves data efficiency and typically performs better in low sample settings, the treatment group in JOBS data had very few positive (unemployment) outcomes. As a result, we observed poor convergence of our MLP models across folds when performing sample splitting on this dataset. Therefore we run JOBS without sample splitting and cross-fitting, and maintain the same setting with OHIE data for consistency.} We use a 4-layer MLP with layer sizes (30, 20, 10) for JOBS data and a 4-layer MLP with layer sizes (40, 30, 10) for OHIE data. We use $\eta=1e-3$ for JOBS data and $\eta=5e-3$ for OHIE data. We train JOBS and OHIE models for 15 and 20 epochs respectively. We with the synthetic experiment, we select hyperparamters by optimizing model performance with respect to the oracle TC model and use the same settings across all models. Note that $\tau = 0.015$ and $\tau = -0.077$ for the outcomes targeted in OHIE and JOBS, respectively. 

\begin{figure}[t]
\includegraphics[width=\linewidth, trim={3mm 3mm 3mm 3mm},clip]{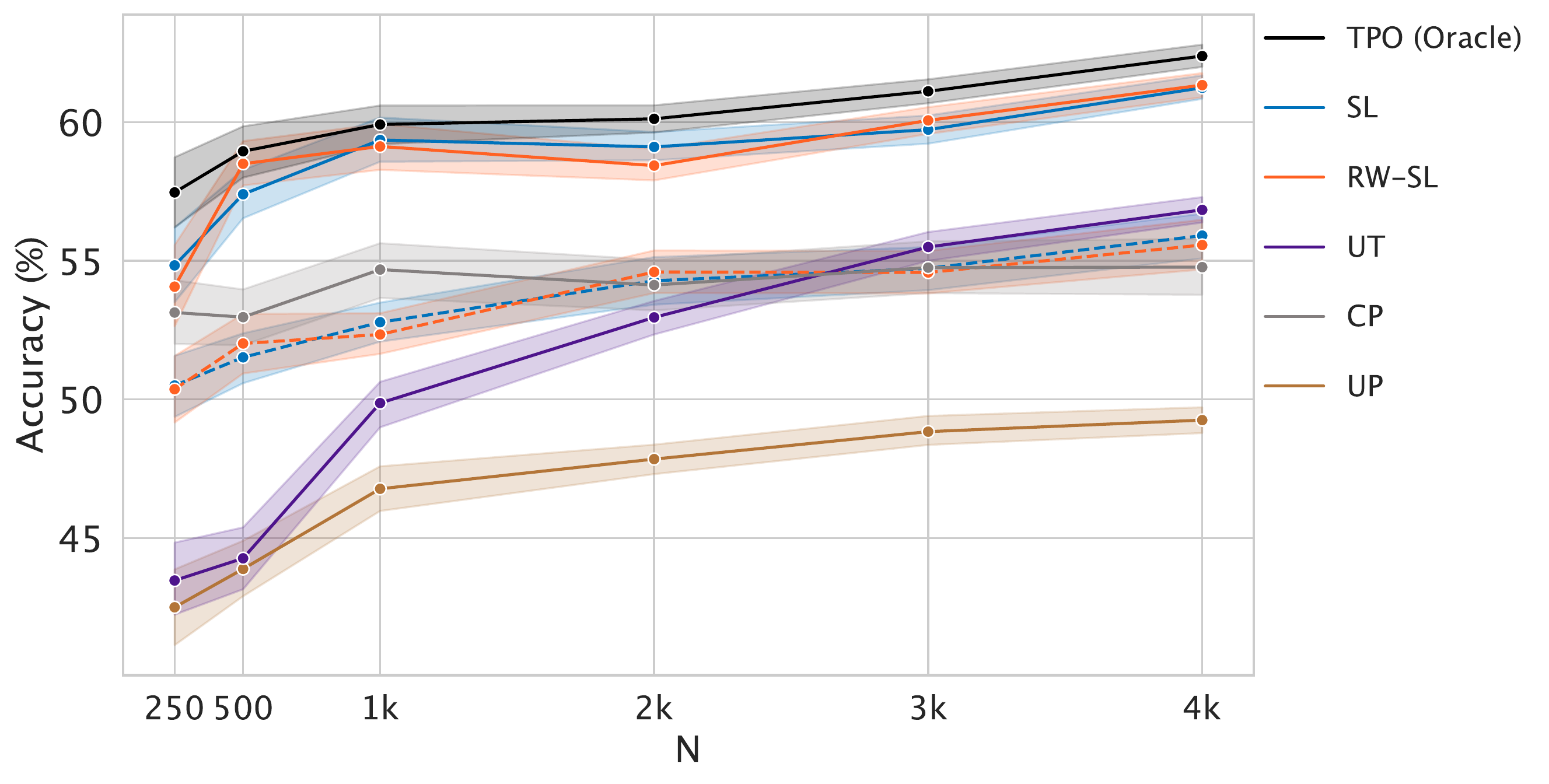}
\caption{Comparison of models across small sample size regimes. SL and RW-SL with oracle parameters maintain performance parity with TPO across settings with 1k+ samples, but demonstrate worse performance than TPO in the most data scarce setting with 250 samples.}\label{fig:syn_convergence_small_n}
\end{figure}

\lgedit{
\subsubsection{Additional results}

Theorem \ref{thm:unbiased_risk} shows that the re-weighted surrogate loss recovers the loss with respect to target potential outcomes in expectation. Because we do not provide a finite sample convergence rate for our method, we extend our synthetic evaluation to a low sample size regime to empirically test the performance of RW-SL on finite samples of limited size. Figure \ref{fig:syn_convergence_small_n} shows a convergence plot for this experiment. We perform this analysis with the same set of hyperparameters used in the main experimental results reported in $\S$ \ref{sec:experiments}. This plot indicates that the performance of all methods deteriorates as sample availability decreases, with performance upper bounded by the oracle TPO model. SL and RW-SL with oracle parameters achieve performance at near parity with TPO in sample settings above 500 samples, and begin to show rapid performance deterioration at 250 samples. This indicates that both SL and RW-SL tend to perform reliably in small sample settings \textit{when parameters and weights are known}. However, SL and RW-SL with learned parameters performs poorly across all sample settings. This is likely due to cascading errors arising from bias in error parameter estimates. UT and UP both learn a function predicting the average outcome response $\hat{\eta}(x) \approx .61, \forall x \in X$ in the setting with 250 and 500 samples. As a result, these methods demonstrate accuracy lower than 50\% in the small sample settings.}

\end{document}